\documentclass{article}
\usepackage{ijcai25}

\usepackage[table,x11names,dvipsnames]{xcolor}         

\usepackage{times}
\usepackage{soul}
\usepackage{url}
\usepackage[hidelinks]{hyperref}
\usepackage[utf8]{inputenc}
\usepackage[small]{caption}
\usepackage{graphicx}
\usepackage{amsmath}
\usepackage{amsthm}
\usepackage{booktabs}
\usepackage{algorithm}
\usepackage{algorithmic}
\usepackage[switch]{lineno}

\title{A-I-RAVEN and I-RAVEN-Mesh: Two New Benchmarks\\for Abstract Visual Reasoning}

\author{
Mikołaj Małkiński$^1$
\And
Jacek Mańdziuk$^{1,2}$\\
\affiliations
$^1$Warsaw University of Technology, Warsaw, Poland\\
$^2$AGH University of Krakow, Krakow, Poland\\
\emails
mikolaj.malkinski.dokt@pw.edu.pl
$\diamond$
jacek.mandziuk@pw.edu.pl
}

\usepackage{amsmath}  
\usepackage{amssymb}
\usepackage{bm}
\usepackage{bbm}
\usepackage{multirow}  
\usepackage{makecell}  
\usepackage{textcomp}  
\usepackage{numprint}  
\usepackage{caption}
\usepackage{subcaption}
\usepackage{float}  
\usepackage[above]{placeins}  
\usepackage{soul}  
\npthousandsep{,}
\usepackage{wrapfig}   
\usepackage{tikz}
\usetikzlibrary{calc,shapes.geometric,trees,positioning,arrows,arrows.meta,backgrounds,fit,decorations.pathreplacing,decorations.markings,calligraphy,matrix}
\usepackage{booktabs}       

\usepackage{setspace}
\usepackage[colaction]{multicol}

\makeatletter
\DeclareRobustCommand{\rvdots}{%
  \vbox{
    \baselineskip2.5\p@\lineskiplimit\z@
    \kern-\p@
    \hbox{.}\hbox{.}\hbox{.}
  }%
}
\makeatother

\begin{document}


\maketitle

\begin{abstract}
We study generalization and knowledge reuse capabilities of deep neural networks in the domain of abstract visual reasoning (AVR), employing Raven's Progressive Matrices (RPMs), a recognized benchmark task for assessing AVR abilities.
Two knowledge transfer scenarios referring to the I-RAVEN dataset are investigated.
Firstly, inspired by generalization assessment capabilities of the PGM dataset and popularity of I-RAVEN, we introduce \emph{Attributeless-I-RAVEN} (A-I-RAVEN), a benchmark with $10$ generalization regimes that allow to systematically test generalization of abstract rules applied to held-out attributes at various levels of complexity (primary and extended regimes).
In contrast to PGM, A-I-RAVEN features compositionality, a variety of figure configurations, and does not require substantial computational resources.
Secondly, we construct \emph{I-RAVEN-Mesh}, a dataset that enriches RPMs with a novel component structure comprising line-based patterns, facilitating assessment of progressive knowledge acquisition in transfer learning setting.
We evaluate 13 strong models from the AVR literature on the introduced datasets, revealing their specific shortcomings in generalization and knowledge transfer.

\end{abstract}

\section{Introduction}
\label{sec:introduction}

\begin{figure}[t]
  \centering
  \includegraphics[width=0.3\textwidth]{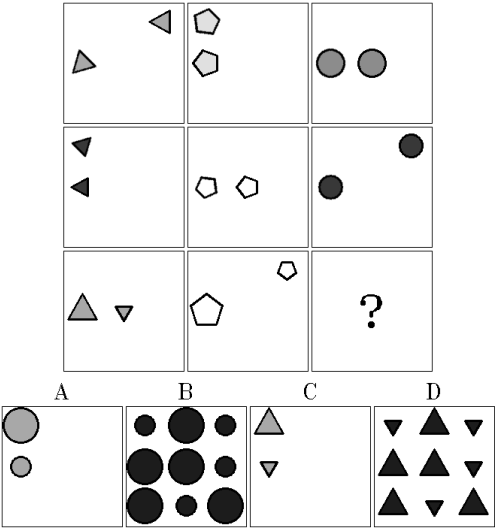}
  \caption{\textbf{RPM example.} The correct answer is A.}
  \label{fig:rpm-example}
\end{figure}
Generalization, the ability of a model to perform well on unseen data, remains a fundamental challenge in deep learning (DL).
While DL methods have demonstrated remarkable achievements in various domains, their generalization capabilities are often questioned, particularly in tasks that demand abstract problem-solving and reasoning skills~\cite{chollet2019measure}.
One such domain is abstract visual reasoning (AVR)~\cite{mitchell2021abstraction,stabinger2021evaluating,van2021much,malkinski2022review} that encompasses tasks requiring (human) fluid intelligence -- an aspect of human cognition believed to be crucial for reasoning in never-encountered settings~\cite{carpenter1990one}.
The most popular AVR tasks are Raven's Progressive Matrices (RPMs)~\cite{raven1936mental,raven1998raven}, which constitute a common problem found in human IQ tests. 
Typical RPMs comprise two components -- the context panels arranged in a $3 \times 3$ grid with the bottom-right panel missing and up to 8 answer panels, out of which only one correctly completes the matrix.
Solving an RPM instance requires identification of underlying abstract rules 
applied to certain attributes 
of the objects composing the instance (see Fig.~\ref{fig:rpm-example}
for an illustrative example).

Design of computational methods capable of tackling RPMs has for decades been an active area of research~\cite{evans1964heuristic,foundalis2006phaeaco,lovett2007analogy,kunda2010taking}.
Consequently, a number of works considered automatic creation of RPM datasets~\cite{matzen2010recreating,wang2015automatic,mandziuk2019deepiq} and a wide suite of predictive models~\cite{hernandez2016computer,hernandez2017measure} were proposed, with DL methods showing the most promising performance~\cite{yang2022conceptual,malkinski2022deep}.
While this rapid progress led to exceeding the human level in particular problem setups~\cite{wu2020scattering,mondal2023learning}, a fundamental challenge of generalization to novel problem settings remains largely unattained.

Initial works designed several  RPM datasets~\cite{matzen2010recreating,wang2015automatic,hoshen2017iq}, however, measuring generalization was not their focus.
While some works explored knowledge transfer between related tasks~\cite{mandziuk2019deepiq,tomaszewska2022duel}, the complexity of the datasets was limited and consequently they didn't pose a challenge for contemporary DL methods.
To measure generalization in modern DL models, the PGM dataset was introduced~\cite{santoro2018measuring}.
PGM defines eight generalization regimes, each specifying the distribution of objects, rules and attributes in train and test splits.
For instance, in the \texttt{Held-out Triples} split, a given rule--object--attribute triplet (e.g. Progression on Object's Size) was assigned only to one of the two splits.
In effect, the models were tested on triplet combinations different from training ones, allowing to assess their generalization capabilities.
A subsequent work proposed RAVEN~\cite{zhang2019raven}, another RPM dataset with enriched perceptual complexity of matrices instantiated in seven visual configurations (\texttt{Center}, \texttt{2x2Grid}, \texttt{3x3Grid}, \texttt{Left-Right}, \texttt{Up-Down}, \texttt{Out-InCenter}, \texttt{Out-InGrid}). 
Moreover, the benchmark is characterized by a moderate sample size, i.e. $70\text{K}$ instances, compared to $1.42\text{M}$ RPMs per each of the eight regimes in PGM.
Due to this size disparity, subsequent research gravitated towards RAVEN and its revised variants (I-RAVEN~\cite{hu2021stratified} and RAVEN-Fair~\cite{benny2020scale}), which didn't require substantial computational resources to train DL models.

\paragraph{Contribution.}
Drawing inspiration from the broad adoption of RAVEN and the generalization assessment capabilities of PGM, this paper proposes a novel suite of generalization challenges stemming from I-RAVEN~\cite{hu2021stratified} (a revised variant of RAVEN that removes a bias in RAVEN's answer panels).
However, unlike I-RAVEN, the proposed suite of benchmarks allows for a direct assessment of the generalization and knowledge transfer of AVR models.
Compared to PGM, our datasets feature compositionality and variety of figure configurations, and their processing doesn't require substantial computational resources.
Furthermore, they include structural annotations, which are utilized, for example, in recent neuro-symbolic approaches~\cite{zhang2021abstract,zhang2022learning}.

First, we introduce \emph{Attributeless-I-RAVEN} (A-I-RAVEN), comprising 10 generalization regimes.
The $4$ primary regimes correspond to specific held-out attributes (\{\texttt{Position}, \texttt{Type}, \texttt{Size}, \texttt{Color}\}), resp.
The training matrices in these regimes adhere to the \texttt{Constant} rule for the respective attribute, whereas test matrices employ a rule different from \texttt{Constant} for this attribute (i.e., \texttt{Progression}, \texttt{Arithmetic}, or \texttt{Distribute Three}).
Moreover, we propose $6$ extended regimes: $3$ of them feature a held-out attribute pair, while another $3$ replace the \texttt{Constant} rule in the training set with each remaining rule.
In effect, each regime comprises different distributions of training and test data.

Next, we propose \emph{I-RAVEN-Mesh}, a variant of I-RAVEN with a new grid-like structure overlaid on the matrices.
The dataset enables assessing generalization to incrementally added structures and progressive knowledge acquisition in a transfer learning (TL) setting.

Investigations involving $13$ contemporary AVR DL models reveal that the introduced benchmarks present a substantial challenge for the tested methods, raising the need for further advancements in this area.

The key contributions of the paper are summarized below.
\begin{itemize}
    \item We introduce the A-I-RAVEN dataset that enables measuring generalization across $10$ regimes.
    \item We construct \textit{I-RAVEN-Mesh}, an extension of I-RAVEN with a new component structure that facilitates assessment of progressive knowledge acquisition in a TL setting.
    \item We evaluate the performance of state-of-the-art AVR models on the introduced benchmarks, uncovering their limitations in terms of generalization to novel problem settings.
\end{itemize}

\section{Related Work}
\label{sec:related-work}

\paragraph{Generalization in AVR.}
In recent years, a variety of AVR problems and corresponding datasets have emerged~\cite{nie2020bongard,fleuret2011comparing,qi2021pqa,shanahan2020explicitly,hill2018learning,zhang2020machine} and several attempts have been made to measure generalization in contemporary AVR models based on the introduced benchmarks.
In particular, distinct visual configurations were employed in RAVEN to assess how a model trained on one configuration performs on the remaining ones~\cite{zhang2019raven,spratley2020closer,zhuo2021effective}.
Although in such a setting the visual aspects of train/test matrices come from different distributions, the underlying rules and attributes remain the same.
In contrast, A-I-RAVEN enables studying the generalization of rules applied to held-out attributes, shifting the focus from perception towards reasoning.
Besides RPMs, the limits of generalization have been explored in other AVR tasks as well.
Visual Analogy Extrapolation Challenge evaluates model's capacity for extrapolation~\cite{webb2020learning}.
However, such specialized datasets might favor models that explicitly embed the notion of extrapolation in their design and aim for being invariant only to specific attributes such as object size or location.
Differently, our benchmarks allow verifying the model's capacity to learn a given concept from the data and generalize it to novel settings.
This perspective links our work to the recent literature on concept learning~\cite{moskvichev2023the}.
However, the concept-oriented benchmarks that originate from ARC~\cite{chollet2019measure} remain largely unsolved by DL models and pose a significant challenge even for leading multi-modal large language models~\cite{mitchell2023comparing}.
In contrast, both benchmarks proposed in this work are attainable by DL models, though further advances in generalization abilities of the models are necessary to consider them solved.

\begin{figure*}[t]
  \centering
  \begin{subfigure}{0.224\textwidth}
    \centering
    \includegraphics[width=\textwidth]{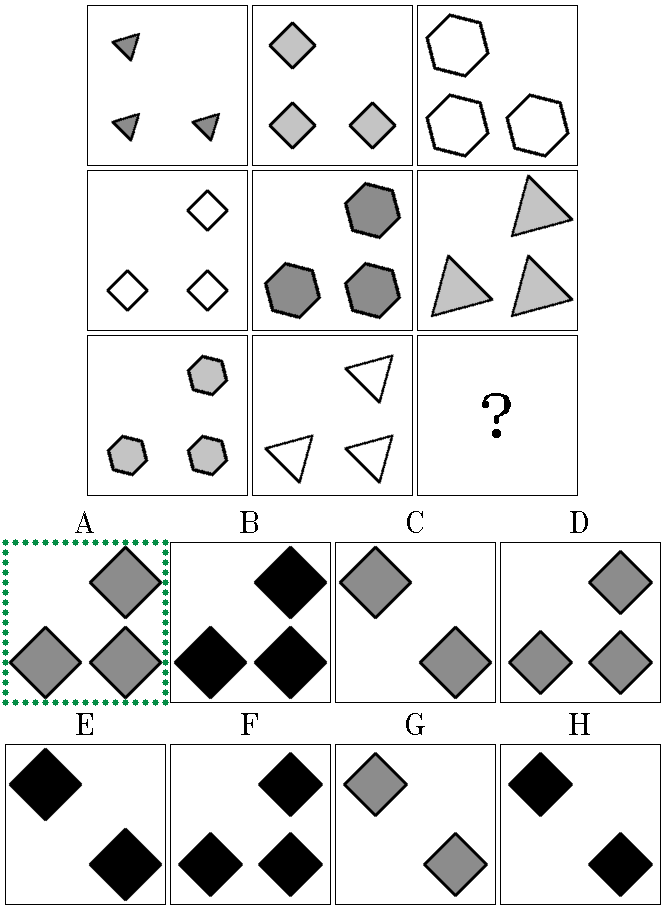}
    \caption{\texttt{A/Position} train}
  \end{subfigure}
  \quad
  \begin{subfigure}{0.224\textwidth}
    \centering
    \includegraphics[width=\textwidth]{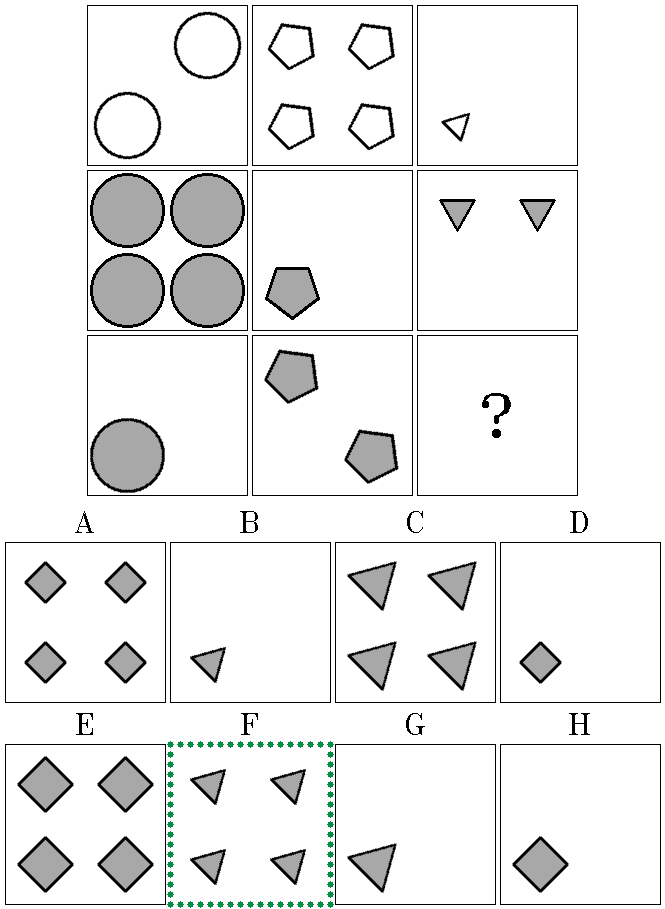}
    \caption{\texttt{A/Position} test}
  \end{subfigure}
  \hfill
  \begin{subfigure}{0.224\textwidth}
    \centering
    \includegraphics[width=\textwidth]{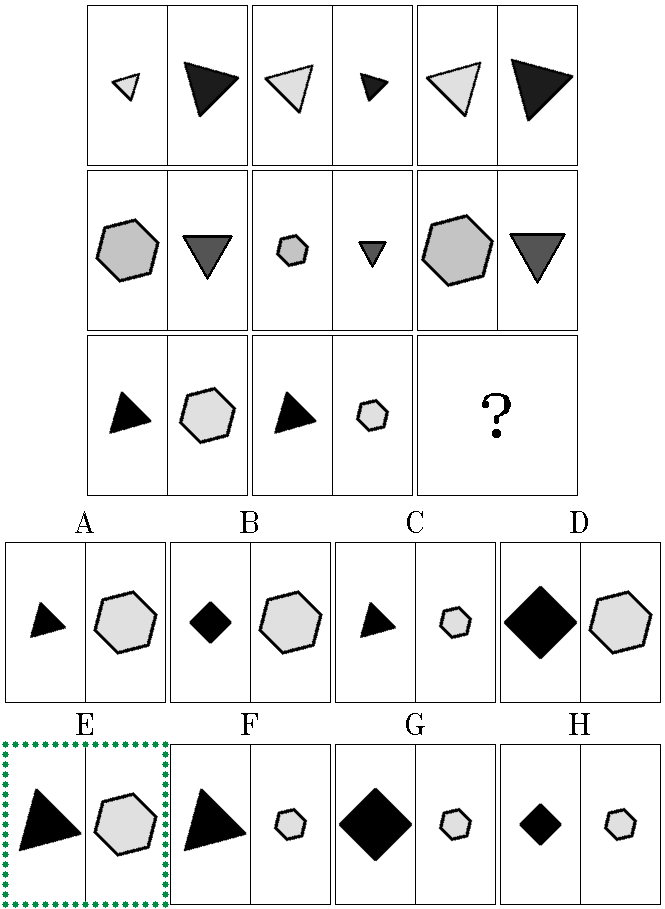}
    \caption{\texttt{A/Color} train}
  \end{subfigure}
  \quad
  \begin{subfigure}{0.224\textwidth}
    \centering
    \includegraphics[width=\textwidth]{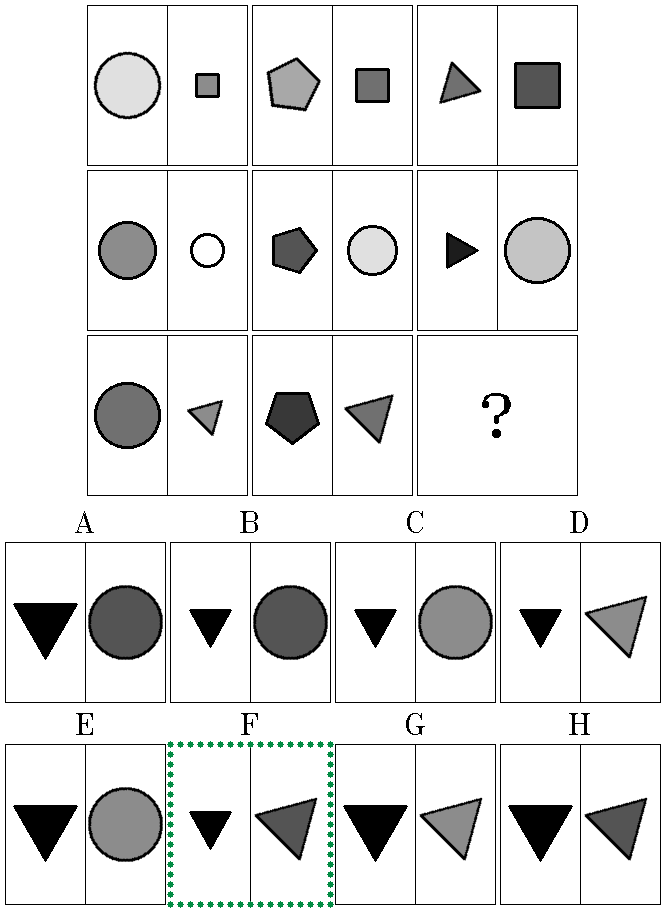}
    \caption{\texttt{A/Color} test}
  \end{subfigure}
  \caption{\textbf{A-I-RAVEN}.
  \underline{Left}:
  Matrices from the \texttt{A/Position} regime belonging to the \texttt{2$\times$2 Grid} configuration.
  In (a), object position is constant across rows, while in (b) object numerosity is governed by \texttt{Distribute Three}.
  \underline{Right}:
  Matrices from the \texttt{A/Color} regime belonging to the \texttt{Left-Right} configuration.
  In (c), object color is constant across rows in left and right image parts, while in (d) it's governed by \texttt{Progression}.
  Correct answers are marked in a green dotted border.
  Please refer to
  Appendix~\ref{sec:examples}
  for examples from other generalization regimes.}
  \label{fig:attributeless-position-color}
\end{figure*}

\begin{figure*}[t]
  \centering
  \begin{subfigure}{0.224\textwidth}
    \centering
    \includegraphics[width=\textwidth]{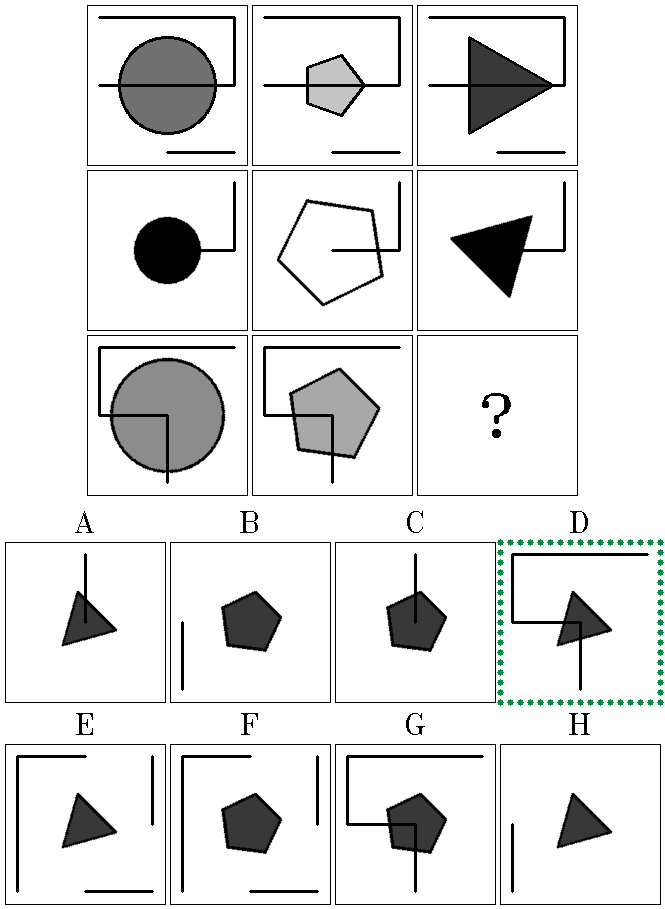}
    \caption{\small\texttt{Constant}}
  \end{subfigure}
  \hfill
  \begin{subfigure}{0.224\textwidth}
    \centering
    \includegraphics[width=\textwidth]{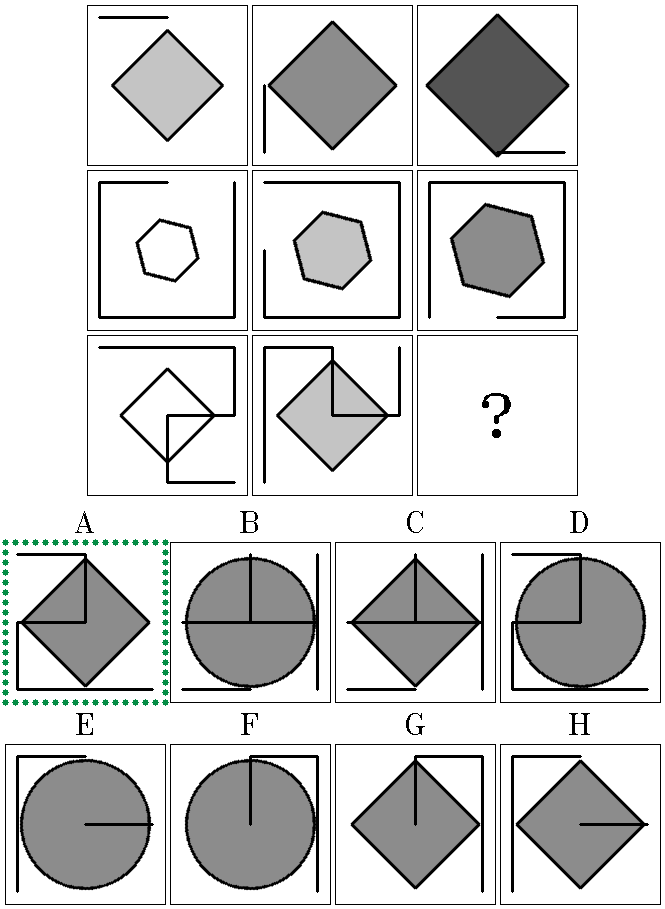}
    \caption{\small\texttt{Progression}}
  \end{subfigure}
  \hfill
  \begin{subfigure}{0.224\textwidth}
    \centering
    \includegraphics[width=\textwidth]{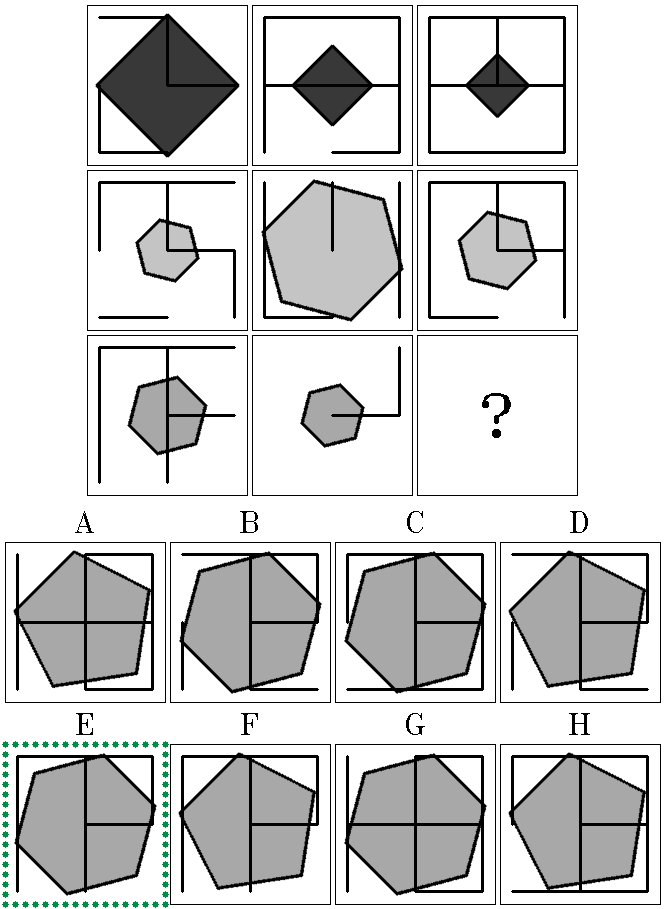}
    \caption{\small\texttt{Arithmetic}}
  \end{subfigure}
  \hfill
  \begin{subfigure}{0.224\textwidth}
    \centering
    \includegraphics[width=\textwidth]{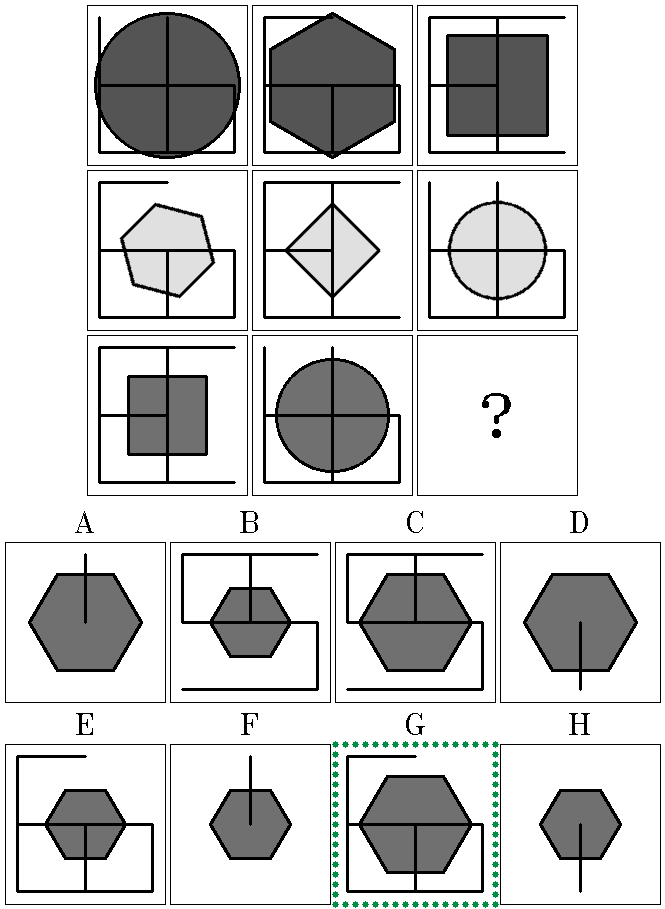}
    \caption{\texttt{Distribute Three}}
  \end{subfigure}
  \caption{\textbf{I-RAVEN-Mesh.}
  Matrices with the \texttt{Position} attribute of the mesh component governed by all applicable rules.
  For the sake of readability, we present examples belonging to the \texttt{Center} configuration.
  (a) Line position is constant in each row.
  (b) The line pattern displayed in the first column is rotated by $90$ degrees in subsequent columns.
  (c) The union set operator applied to the first and the second column produces line positions in the third column.
  (d) Each row contains lines arranged in one out of three available patterns.
  Correct answers are marked in a green dotted border.
  Please refer to
  Appendix~\ref{sec:examples}
  for examples concerning the \texttt{Number} attribute.}
  \label{fig:mesh-position}
\end{figure*}

\paragraph{Model architectures.}
Preliminary attempts to solve RPMs with DL models involve WReN~\cite{santoro2018measuring} that reasons over object relations using Relation Network~\cite{santoro2017simple}, or SRAN~\cite{hu2021stratified} that relies on a hierarchical architecture with panel encoders devoted to particular image groups.
A common theme enabling generalization in DL models is to explicitly identify RPM objects.
To this end, RelBase~\cite{spratley2020closer} employs Attend-Infer-Repeat, an unsupervised scene decomposition method, STSN~\cite{mondal2023learning} utilizes Slot attention~\cite{locatello2020object} to decompose matrix to slots containing particular objects and  Temporal Context Normalization (TCN)~\cite{webb2020learning} to normalize latent matrix panel representations in a task-specific context, DRNet~\cite{zhao2024learning} relies on a dual-stream design,
and MRNet~\cite{benny2020scale} presents a multi-scale architecture.
SCL~\cite{wu2020scattering} proposes the scattering transformation, CoPINet~\cite{zhang2019learning} and CPCNet~\cite{yang2023cognitively} rely on contrastive architectures, PredRNet~\cite{yang2023neural} learns to minimize the prediction error, ALANS~\cite{zhang2021abstract} and PrAE~\cite{zhang2022learning} employ neuro-symbolic architectures, and SCAR~\cite{malkinski2024one} adapts its computation to the structure of the considered matrix.
Despite the high variety of AVR models, experiments on the introduced benchmarks reveal their shortcomings in terms of generalization and knowledge transfer.

\section{Proposed datasets}
\label{sec:method}

The set of attributes in I-RAVEN is $\mathcal{A}=\{$\texttt{Position}, \texttt{Number}, \texttt{Type}, \texttt{Size}, \texttt{Color}$\}$ and the set of rules is $\mathcal{R}=\{$\texttt{Constant}, \texttt{Progression}, \texttt{Arithmetic}, \texttt{Distribute Three}$\}$.
For attribute $a \in \mathcal{A}$ and a dataset split $s \in \mathcal{S}$, where $\mathcal{S}=\{$train, val., test$\}$, we define the set of rules applicable to $a$ in split $s$ by $R(a,s) \subseteq \mathcal{R}$.
In I-RAVEN all rule--attribute pairs are valid in all splits:
\begin{equation}
    R(a,s) = \mathcal{R}, \quad \forall a \in \mathcal{A} \land \forall s \in \mathcal{S}
\end{equation}

\subsection{Attributeless-I-RAVEN (A-I-RAVEN)}
To probe generalization in DL models, we present A-I-RAVEN, a benchmark composed of $10$ generalization regimes.
Example matrices are illustrated in Fig.~\ref{fig:attributeless-position-color}, with additional samples provided in
Appendix~\ref{sec:examples}.
Each regime defines a set of held-out attributes $A^*$, each with a corresponding rule $r^*(a), a \in A^*$.
In train and validation splits, held-out attribute $a \in A^*$ is governed by $r^*(a)$.
In the test split, $a \in A^*$ is governed by a different rule sampled from $\mathcal{R}-\{r^*(a)\}$.
In effect, during training, the model doesn't see rule--attribute combinations required to solve test matrices.
There are no rule-related constraints on the remaining attributes.
In summary, we have:
\begin{equation}
    R(a,s) = \begin{cases}
        \{r^*(a)\} & \text{if } a \in A^* \land s \in \{\text{train}, \text{val.}\}, \\
        \mathcal{R} - \{r^*(a)\} & \text{if } a \in A^* \land s = \text{test}, \\
        \mathcal{R} & \text{if } a \not\in A^*. \\
    \end{cases}
\end{equation}
We define $4$ primary regimes with $r^*(a) = \texttt{Constant}$ that correspond to
individual held-out attributes ($\vert A^* \vert = 1$), denoted as \texttt{A/<Attribute>} (e.g., \texttt{A/Type}).
Since \texttt{Position} and \texttt{Number} attributes are tightly coupled (e.g., it's impossible to increase cardinality of objects while keeping their position constant), we allocate a single generalization regime, \texttt{A/Position}, to cover both attributes.
In addition, we define $6$ extended regimes as supplementary generalization challenges.
In the first group a pair of attributes is held-out in the training set, i.e. $\vert A^* \vert = 2$.
Specifically, we introduce $3$ new regimes: \texttt{A/ColorSize}, \texttt{A/ColorType}, and \texttt{A/SizeType}, based on the respective attribute pairs.
In the second group, \texttt{Constant} rule in $r^*(a)$ is replaced with each of the $3$ remaining rules, leading to \texttt{A/Color-Progression}, \texttt{A/Color-Arithmetic}, and \texttt{A/Color-DistributeThree} regimes.
While this modification could be applied to all the described regimes, we focus on the \texttt{Color} attribute due to its broad range of possible values.

\begin{table*}[t]
    \centering
    \small
    \begin{tabular}{p{0.08\textwidth} p{0.175\textwidth} p{0.67\textwidth}}
        \toprule
        Attribute & Rule & Description \\
        \midrule
        \multirow{5}{*}{\texttt{Number}} & \texttt{Constant} & Each image in a given row contains the same number of lines. \\
        & \texttt{Progression} & The count of lines in a given row changes by a constant factor (e.g. $2, 4, 6$). \\
        & \texttt{Arithmetic} & The number of lines in the third column is determined based on an arithmetic operation applied to the preceding columns (e.g. $3 - 1 = 2$). \\
        & \texttt{Distribute Three} & Three line counts are sampled and spread among images in a given row. \\
        \midrule
        \multirow{5}{*}{\texttt{Position}} & \texttt{Constant} & Each image in a given row contains the same position of lines. \\
        & \texttt{Progression} & A panel arrangement is sampled in each row and rotated by $90$ degrees in subsequent columns. \\
        & \texttt{Arithmetic} & The position of lines in the third column is computed based on a set operation (union or difference) applied to the preceding columns. \\
        & \texttt{Distribute Three} & Three line arrangements are sampled and spread among images in a given row. \\
        \bottomrule
    \end{tabular}
    \caption{Description of rule--attribute pairs in I-RAVEN-Mesh.}
    \label{tab:mesh-description}
\end{table*}

\subsection{I-RAVEN-Mesh}
\label{sec:i-raven-mesh}
The other of the proposed benchmarks is designed to probe progressive knowledge acquisition in a TL setting.
I-RAVEN-Mesh extends I-RAVEN by introducing a novel visual component overlaid on top of the existing I-RAVEN components (see Fig.~\ref{fig:mesh-position}).
Though the dataset can serve as a learning challenge on its own, the main motivation behind its introduction is to employ models pre-trained on I-RAVEN and fine-tune them on I-RAVEN-Mesh with a configurable train sample size, facilitating analysis of their TL performance.
The mesh grid comprises from $1$ to $12$ lines placed in predefined locations.
The set of available lines covers the inner and outer edges of a $2 \times 2$ grid ($12$ lines in total).
The mesh component has two attributes: $\mathcal{A}^{\text{mesh}}=\{\texttt{Number}, \texttt{Position}\}$, which govern the count and location of lines, respectively.
To each attribute a rule $r \in \mathcal{R}$ can be applied.
Table~\ref{tab:mesh-description} describes the effect of applying a given rule--attribute pair to the mesh component.
To generate the mesh component of an I-RAVEN-Mesh matrix, we sample one of the two attributes $a \in \mathcal{A}^{\text{mesh}}$ and a corresponding rule $r \in \mathcal{R}$ that governs its values.
As the attributes often depend on each other (e.g., it's impossible to increase the number of lines while keeping their position constant), we don't constrain the value of the other attribute.
The rule--attribute pairs for the base I-RAVEN components are generated in the same way as in the original dataset.
To generate answers to the matrix, we follow the impartial algorithm proposed in I-RAVEN~\cite{hu2021stratified}.
In addition, each matrix contains at least one incorrect answer that differs from the correct one only in the mesh component, ensuring that the solver has to identify the correct rule governing the mesh component in order to solve the matrix.
To facilitate training with an auxiliary loss, in which the model additionally predicts the representation of rules governing the matrix~\cite{santoro2018measuring}, we extend the base set of rule annotations with ones concerning the Mesh component.

\begin{table*}[t]
\centering\small
\begin{tabular}{l|c|c|c|cccc}
\toprule
 & I-RAVEN$^\dagger$ & I-RAVEN (ours) & Mesh & \texttt{A/Color} & \texttt{A/Position} & \texttt{A/Size} & \texttt{A/Type} \\
\midrule
ALANS & $-$ & $27.0$ \scriptsize$(\pm\,8.4)$ & $15.9$ \scriptsize$(\pm\,2.6)$ & $15.2$ \scriptsize$(\pm\,1.4)$ & $16.0$ \scriptsize$(\pm\,1.0)$ & $23.3$ \scriptsize$(\pm\,6.5)$ & $19.0$ \scriptsize$(\pm\,3.4)$ \\
CPCNet & $\textbf{98.5}$ & $70.4$ \scriptsize$(\pm\,6.4)$ & $66.6$ \scriptsize$(\pm\,5.1)$ & $51.2$ \scriptsize$(\pm\,3.8)$ & $68.3$ \scriptsize$(\pm\,4.0)$ & $43.5$ \scriptsize$(\pm\,3.5)$ & $38.6$ \scriptsize$(\pm\,4.3)$ \\
CNN-LSTM & $18.9$ & $27.5$ \scriptsize$(\pm\,1.5)$ & $28.9$ \scriptsize$(\pm\,0.4)$ & $17.0$ \scriptsize$(\pm\,3.1)$ & $24.0$ \scriptsize$(\pm\,2.9)$ & $13.6$ \scriptsize$(\pm\,1.4)$ & $14.5$ \scriptsize$(\pm\,0.8)$ \\
CoPINet & $46.1$ & $43.2$ \scriptsize$(\pm\,0.1)$ & $41.1$ \scriptsize$(\pm\,0.3)$ & $32.5$ \scriptsize$(\pm\,0.2)$ & $41.3$ \scriptsize$(\pm\,1.6)$ & $21.8$ \scriptsize$(\pm\,0.2)$ & $19.8$ \scriptsize$(\pm\,0.9)$ \\
DRNet & $\underline{97.6}$ & $\textbf{90.9}$ \scriptsize$(\pm\,1.1)$ & $\underline{83.9}$ \scriptsize$(\pm\,2.7)$ & $\textbf{70.0}$ \scriptsize$(\pm\,1.6)$ & $\textbf{77.5}$ \scriptsize$(\pm\,0.9)$ & $\underline{54.3}$ \scriptsize$(\pm\,3.0)$ & $\underline{44.3}$ \scriptsize$(\pm\,0.8)$ \\
MRNet & $83.5$ & $86.7$ \scriptsize$(\pm\,2.3)$ & $79.5$ \scriptsize$(\pm\,2.0)$ & $33.6$ \scriptsize$(\pm\,8.2)$ & $62.6$ \scriptsize$(\pm\,2.6)$ & $20.6$ \scriptsize$(\pm\,5.0)$ & $19.4$ \scriptsize$(\pm\,0.3)$ \\
PrAE & $77.0$ & $19.5$ \scriptsize$(\pm\,0.4)$ & $33.2$ \scriptsize$(\pm\,0.4)$ & $47.9$ \scriptsize$(\pm\,0.9)$ & $68.2$ \scriptsize$(\pm\,3.3)$ & $41.3$ \scriptsize$(\pm\,1.8)$ & $37.0$ \scriptsize$(\pm\,1.7)$ \\
PredRNet & $96.5$ & $88.8$ \scriptsize$(\pm\,1.8)$ & $59.2$ \scriptsize$(\pm\,6.4)$ & $59.4$ \scriptsize$(\pm\,1.0)$ & $73.7$ \scriptsize$(\pm\,0.7)$ & $47.5$ \scriptsize$(\pm\,1.3)$ & $40.2$ \scriptsize$(\pm\,1.3)$ \\
RelBase & $91.1$ & $\underline{89.6}$ \scriptsize$(\pm\,0.6)$ & $\textbf{84.9}$ \scriptsize$(\pm\,4.4)$ & $\underline{67.4}$ \scriptsize$(\pm\,2.7)$ & $76.6$ \scriptsize$(\pm\,0.3)$ & $51.1$ \scriptsize$(\pm\,2.4)$ & $44.1$ \scriptsize$(\pm\,1.0)$ \\
SCL & $95.0$ & $83.4$ \scriptsize$(\pm\,2.5)$ & $80.9$ \scriptsize$(\pm\,1.5)$ & $65.1$ \scriptsize$(\pm\,2.0)$ & $\underline{76.7}$ \scriptsize$(\pm\,7.1)$ & $\textbf{65.6}$ \scriptsize$(\pm\,2.4)$ & $\textbf{49.5}$ \scriptsize$(\pm\,1.8)$ \\
SRAN & $60.8$ & $58.2$ \scriptsize$(\pm\,1.6)$ & $57.8$ \scriptsize$(\pm\,0.2)$ & $38.3$ \scriptsize$(\pm\,1.0)$ & $56.9$ \scriptsize$(\pm\,0.7)$ & $34.4$ \scriptsize$(\pm\,3.0)$ & $30.7$ \scriptsize$(\pm\,2.2)$ \\
STSN & $95.7$ & $59.0$ \scriptsize$(\pm\,18.5)$ & $48.7$ \scriptsize$(\pm\,11.5)$ & $39.3$ \scriptsize$(\pm\,6.9)$ & $36.1$ \scriptsize$(\pm\,19.9)$ & $38.4$ \scriptsize$(\pm\,16.6)$ & $39.1$ \scriptsize$(\pm\,5.0)$ \\
WReN & $23.8$ & $18.4$ \scriptsize$(\pm\,0.0)$ & $25.7$ \scriptsize$(\pm\,0.2)$ & $16.9$ \scriptsize$(\pm\,0.5)$ & $17.3$ \scriptsize$(\pm\,0.4)$ & $12.4$ \scriptsize$(\pm\,0.5)$ & $15.1$ \scriptsize$(\pm\,0.7)$ \\
\bottomrule
\end{tabular}
\caption{\textbf{Single-task learning.}
    Mean and standard deviation of test accuracy for three random seeds. 
    Best dataset results are marked in bold and the second best are underlined.
    I-RAVEN$^\dagger$ provides results on I-RAVEN reported by model authors in the corresponding papers, while I-RAVEN (ours) presents results obtained with our experimental setup, which utilizes a typical configuration of an optimizer and learning rate scheduler without model-specific tuning, and doesn't involve data augmentation, see "Experimental setup" in Section~\ref{sec:experiments} for details.}
\label{tab:attributeless}
\end{table*}
\begin{table*}[t]
\centering\small
\begin{tabular}{l|cccccc}
\toprule
 & \texttt{A/ColorSize} & \texttt{A/ColorType} & \texttt{A/SizeType} & \texttt{A/Color-P} & \texttt{A/Color-A} & \texttt{A/Color-D3} \\
\midrule
ALANS & $15.1$ \scriptsize$(\pm\,3.3)$ & $17.7$ \scriptsize$(\pm\,3.2)$ & $15.7$ \scriptsize$(\pm\,3.2)$ & $24.8$ \scriptsize$(\pm\,18.8)$ & $18.3$ \scriptsize$(\pm\,6.6)$ & $22.4$ \scriptsize$(\pm\,7.7)$ \\
CPCNet & $33.0$ \scriptsize$(\pm\,5.3)$ & $25.0$ \scriptsize$(\pm\,0.9)$ & $24.1$ \scriptsize$(\pm\,1.2)$ & $50.5$ \scriptsize$(\pm\,0.6)$ & $45.9$ \scriptsize$(\pm\,2.7)$ & $37.8$ \scriptsize$(\pm\,0.9)$ \\
CNN-LSTM & $13.4$ \scriptsize$(\pm\,0.9)$ & $14.7$ \scriptsize$(\pm\,1.7)$ & $13.0$ \scriptsize$(\pm\,0.1)$ & $17.2$ \scriptsize$(\pm\,1.5)$ & $17.1$ \scriptsize$(\pm\,3.7)$ & $20.6$ \scriptsize$(\pm\,6.7)$ \\
CoPINet & $18.3$ \scriptsize$(\pm\,0.3)$ & $17.2$ \scriptsize$(\pm\,0.1)$ & $19.7$ \scriptsize$(\pm\,0.7)$ & $35.8$ \scriptsize$(\pm\,0.6)$ & $35.2$ \scriptsize$(\pm\,0.5)$ & $26.9$ \scriptsize$(\pm\,0.5)$ \\
DRNet & $\underline{38.3}$ \scriptsize$(\pm\,0.5)$ & $29.5$ \scriptsize$(\pm\,0.5)$ & $\underline{31.6}$ \scriptsize$(\pm\,1.2)$ & $72.8$ \scriptsize$(\pm\,1.3)$ & $\textbf{66.7}$ \scriptsize$(\pm\,1.2)$ & $63.2$ \scriptsize$(\pm\,0.3)$ \\
MRNet & $18.7$ \scriptsize$(\pm\,1.1)$ & $20.0$ \scriptsize$(\pm\,2.6)$ & $28.2$ \scriptsize$(\pm\,0.9)$ & $34.4$ \scriptsize$(\pm\,3.4)$ & $35.7$ \scriptsize$(\pm\,5.9)$ & $18.6$ \scriptsize$(\pm\,0.1)$ \\
PrAE & $30.0$ \scriptsize$(\pm\,1.1)$ & $26.7$ \scriptsize$(\pm\,0.7)$ & $25.6$ \scriptsize$(\pm\,0.8)$ & $62.3$ \scriptsize$(\pm\,0.9)$ & $43.0$ \scriptsize$(\pm\,26.5)$ & $55.1$ \scriptsize$(\pm\,0.8)$ \\
PredRNet & $31.0$ \scriptsize$(\pm\,1.6)$ & $28.0$ \scriptsize$(\pm\,0.7)$ & $27.9$ \scriptsize$(\pm\,0.5)$ & $62.3$ \scriptsize$(\pm\,2.2)$ & $56.9$ \scriptsize$(\pm\,1.4)$ & $48.5$ \scriptsize$(\pm\,0.9)$ \\
RelBase & $36.6$ \scriptsize$(\pm\,0.8)$ & $\underline{29.7}$ \scriptsize$(\pm\,0.6)$ & $31.1$ \scriptsize$(\pm\,1.0)$ & $\underline{73.0}$ \scriptsize$(\pm\,1.8)$ & $\underline{66.2}$ \scriptsize$(\pm\,1.0)$ & $\textbf{65.7}$ \scriptsize$(\pm\,4.6)$ \\
SCL & $\textbf{40.8}$ \scriptsize$(\pm\,3.2)$ & $\textbf{32.0}$ \scriptsize$(\pm\,2.3)$ & $\textbf{33.5}$ \scriptsize$(\pm\,0.7)$ & $\textbf{75.6}$ \scriptsize$(\pm\,10.1)$ & $60.0$ \scriptsize$(\pm\,4.1)$ & $\underline{63.9}$ \scriptsize$(\pm\,4.3)$ \\
SRAN & $22.7$ \scriptsize$(\pm\,1.1)$ & $20.9$ \scriptsize$(\pm\,0.9)$ & $23.3$ \scriptsize$(\pm\,0.3)$ & $42.1$ \scriptsize$(\pm\,2.3)$ & $39.9$ \scriptsize$(\pm\,2.7)$ & $34.6$ \scriptsize$(\pm\,3.6)$ \\
STSN & $27.3$ \scriptsize$(\pm\,4.6)$ & $21.9$ \scriptsize$(\pm\,4.6)$ & $12.3$ \scriptsize$(\pm\,0.1)$ & $39.9$ \scriptsize$(\pm\,14.7)$ & $25.7$ \scriptsize$(\pm\,10.6)$ & $20.7$ \scriptsize$(\pm\,7.7)$ \\
WReN & $13.5$ \scriptsize$(\pm\,0.1)$ & $13.8$ \scriptsize$(\pm\,0.7)$ & $14.1$ \scriptsize$(\pm\,0.2)$ & $18.0$ \scriptsize$(\pm\,0.4)$ & $17.1$ \scriptsize$(\pm\,0.2)$ & $17.7$ \scriptsize$(\pm\,0.6)$ \\
\bottomrule
\end{tabular}
\caption{\textbf{A-I-RAVEN extended regimes.} P, A, and D3 denote Progression, Arithmetic, and Distribute Three, resp.}
\label{tab:attributeless-extended}
\end{table*}

\section{Experiments}
\label{sec:experiments}

We assess generalization of state-of-the-art models for solving RPMs on A-I-RAVEN and evaluate progressive knowledge acquisition on I-RAVEN-Mesh.

\paragraph{Experimental setup.}
In all experiments we use the Adam optimizer~\cite{kingma2014adam} with $\beta_1=0.9$, $\beta_2=0.999$, $\epsilon=10^{-8}$ and a batch size set to $128$.
Learning rate is initialized to $0.001$ and reduced $10$-fold (at most 3 times) if no progress is seen in the validation loss in 5 subsequent epochs, and training stops early in the case of $10$ epochs without progress.
Unless stated otherwise, each model configuration was trained $3$ times with a different seed, and we report mean and standard deviation for these runs.
In each experiment, we utilize $42\,000$ training, $14\,000$ validation, and $14\,000$ test matrices, following the standard data split protocol taken in prior works~\cite{zhang2019raven,hu2021stratified}.
All models are trained with the auxiliary loss with sparse encoding~\cite{malkinski2020multi} and $\beta=1$.
Experiments were run on a worker with a single NVIDIA DGX A100 GPU.

\paragraph{Models.}
In addition to the simple CNN-LSTM baseline~\cite{santoro2018measuring}, we assess generalization of SOTA AVR models including WReN~\cite{santoro2018measuring}, CoPINet~\cite{zhang2019learning}, RelBase~\cite{spratley2020closer}, SCL~\cite{wu2020scattering}, MRNet~\cite{benny2020scale}, ALANS~\cite{zhang2021abstract}, SRAN~\cite{hu2021stratified}, PrAE~\cite{zhang2022learning}, CPCNet~\cite{yang2023cognitively}, PredRNet~\cite{yang2023neural}, STSN~\cite{mondal2023learning}, and DRNet~\cite{zhao2024learning}.
For direct comparison, we evaluate all models on I-RAVEN following the above-described experimental setup.

\paragraph{Reproducibility.}
To guarantee reproducibility of experiments, we use a fixed set of random seeds and turn off hardware and framework features concerning indeterministic computation wherever possible.
Together with the code, we provide the full training script that can be used to run all training jobs.
The training job is packaged as a Docker image with fixed dependencies to isolate the configuration of the training environment.
The released code allows for generation of all datasets from scratch, eliminating the dependency on file-hosting services required to distribute the data.
The code for reproducing all experiments is publicly accessible at: \url{https://github.com/mikomel/raven}

\subsection{Generalization on A-I-RAVEN}
\label{sec:experiments-generalization}
\paragraph{Main regimes.}
In the first set of experiments we evaluate all considered models on $4$ primary generalization regimes of A-I-RAVEN.
The results are presented in Table~\ref{tab:attributeless}, along with the reference results on I-RAVEN and I-RAVEN-Mesh.
The best outcomes on \texttt{A/Color} and \texttt{A/Position} are achieved by DRNet, followed by RelBase and SCL that perform comparably.
In \texttt{A/Size} and \texttt{A/Type}, SCL outperforms other models, with DRNet and RelBase taking the second and third place, resp.
Interestingly, the top 3 models present a mix of architectures comprising large models, such as DRNet that includes a Vision Transformer backbone ($24.7$M params), as well as small models, such as SCL and RelBase that include mainly convolutional and feed-forward layers ($0.6$M and $1.3$M params, resp.).
This suggests that various architectural approaches may be taken to achieve reasonable generalization performance in solving RPMs.

\paragraph{Extended regimes.}
Table~\ref{tab:attributeless-extended} shows the aggregated performance of all considered models on $6$ extended A-I-RAVEN regimes.
Similarly to the main regimes, the best results are achieved by SCL, RelBase, and DRNet.
Overall, replacing the \texttt{Constant} rule in the training set of the \texttt{A/Color} regime with \texttt{Progression} yields a dataset of slightly lower complexity, as the best model on \texttt{A/Color-Progression} achieved $75.6\%$ accuracy, a $5.6$ p.p. increase compared to the best result on \texttt{A/Color}.
Conversely, using the \texttt{Arithmetic} and \texttt{Distribute Three} rules increases the difficulty, as measured by the drop of the max accuracy by $3.3$ p.p. and $4.3$ p.p., resp.
Furthermore, using a pair of held-out attributes significantly increases the complexity.
For instance, in \texttt{A/ColorType}, the most challenging regime, the best result is only $32.0\%$.
We conclude that A-I-RAVEN provides a suite of challenging regimes of variable complexity, in which even the best-performing models are far from solving all test matrices.

\begin{figure}[t]
  \centering
  \includegraphics[width=0.47\textwidth]{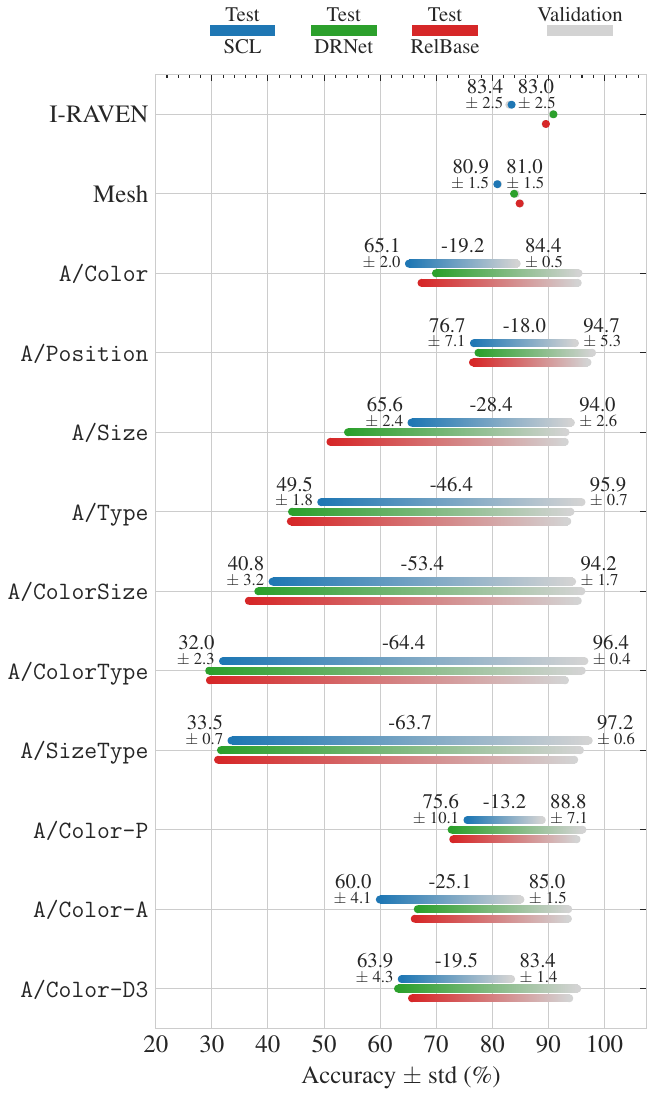}
  \caption{\textbf{Dataset difficulty.}
  Performance of top-3 models on test and validation splits.
  Numerical values refer to SCL scores.
  }
  \label{fig:difficulty}
\end{figure}

\paragraph{Dataset difficulty.}
Across all A-I-RAVEN regimes, the highest average result was achieved by SCL ($56.3\%$), followed by DRNet ($54.8\%$) and RelBase ($54.1\%$).
While SCL achieved $83.4\%$ test accuracy on I-RAVEN, on A-I-RAVEN regimes it scored from $32.0\%$ on \texttt{A/ColorType} to $76.7\%$ on \texttt{A/Position}.
Similar differences can be observed for all remaining models, which shows that generalization regimes of A-I-RAVEN pose a bigger challenge than the base dataset.

Fig.~\ref{fig:difficulty} displays the difference in performance of top-3 models on test and validation splits.
On I-RAVEN and I-RAVEN-Mesh the difference is negligible, as in these datasets both splits follow the same distribution.
However, the difference in attributeless regimes is significant, indicating the need for further research on generalization.

Tables~\ref{tab:stl-aux-i-raven} --~\ref{tab:stl-aux-color-distributethree} in Appendix~\ref{sec:extended-results}
present the results of all considered models on test and validation splits and the difference between these two splits for particular datasets/regimes.
The difference in model performance between test and validation splits in I-RAVEN
(Table~\ref{tab:stl-aux-i-raven})
and I-RAVEN-Mesh
(Table~\ref{tab:stl-aux-mesh})
is negligible.
In A-I-RAVEN regimes, however, the difference is significant, showing limitations of all evaluated models in terms of generalization.
Across $4$ primary regimes
(Tables~\ref{tab:stl-aux-color} --~\ref{tab:stl-aux-type}),
the biggest difference concerns the \texttt{A/Type} regime, suggesting that generalization of rules applied to novel shape types constitutes a real challenge for the contemporary models.
In all $3$ extended regimes concerning held-out attribute pairs (\texttt{A/ColorSize}, \texttt{A/ColorType}, and \texttt{A/SizeType}) the performance difference on test and validation splits is bigger than in the primary regimes
(see Tables~\ref{tab:stl-aux-color-size} --~\ref{tab:stl-aux-size-type}).
This drop stems from overall weaker performance on the test split, confirming high difficulty of these regimes.
Model performance on the next $3$ regimes concerning the \texttt{Color} attribute and rules other than \texttt{Constant} (\texttt{A/Color-Progression}, \texttt{A/Color-Arithmetic}, and \texttt{A/Color-DistributeThree}) is better, though further progress in generalization is required to fully close the performance gap between test and validation splits
(see Tables~\ref{tab:stl-aux-color-progression} --~\ref{tab:stl-aux-color-distributethree}).

\begin{figure*}[t]
  \centering
  \includegraphics[width=0.98\textwidth]{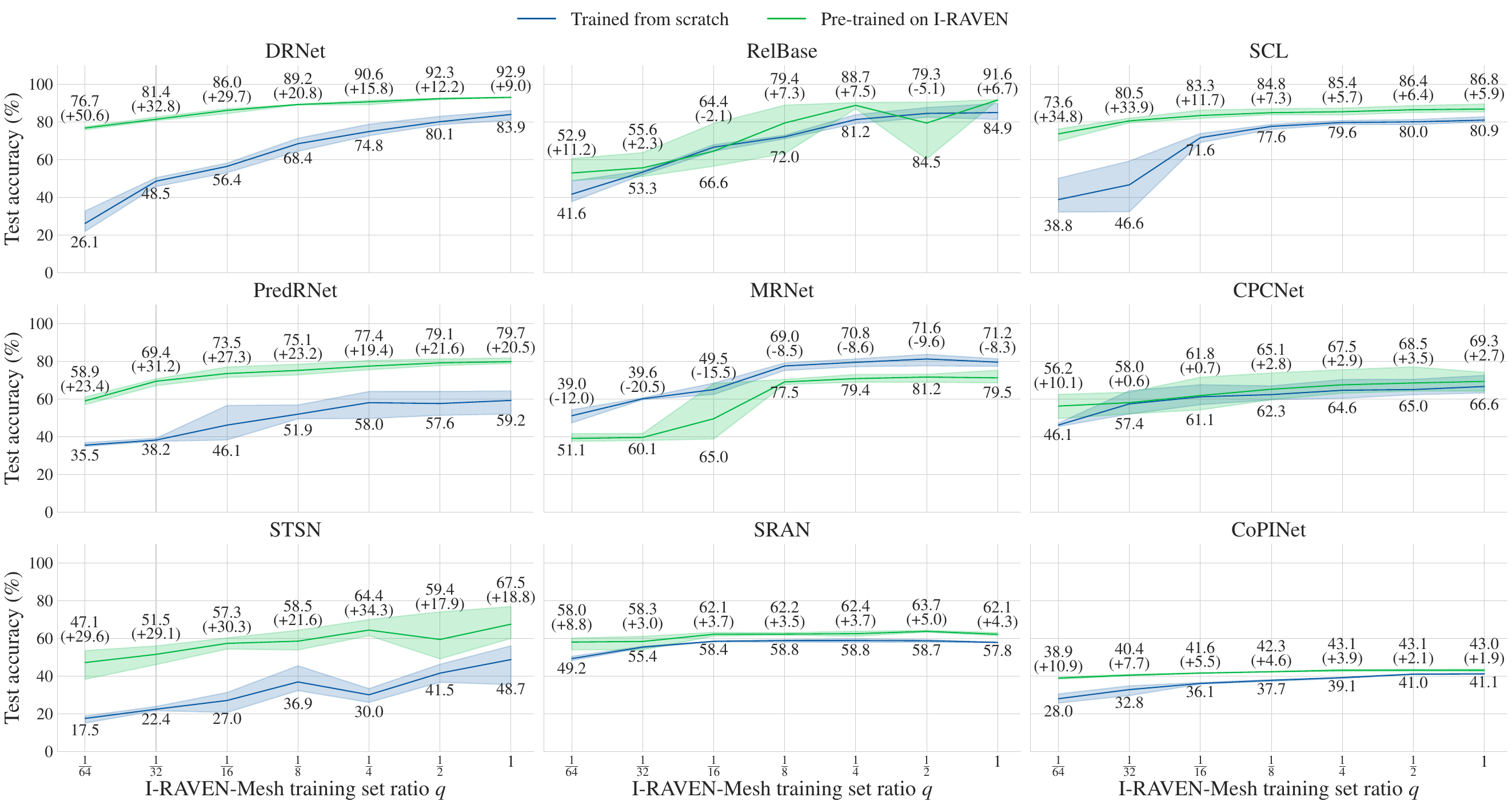}
  \caption{\textbf{Transfer learning.}
  Mean and standard deviation of test accuracy on I-RAVEN-Mesh across three random seeds.
  Models were trained in two setups:
  1) from scratch on I-RAVEN-Mesh with variable sample size;
  2) pre-trained on full I-RAVEN and fine-tuned on I-RAVEN-Mesh with variable sample size.
  Results for setups 1) and 2) are shown below and above the plot lines, resp.
  }
  \label{fig:transfer-learning}
\end{figure*}

\paragraph{Per-configuration results.}
Tables~\ref{tab:stl-configurations-i-raven} --~\ref{tab:stl-configurations-color-distributethree} in Appendix~\ref{sec:extended-results}
present the detailed results of all considered models for all matrix configurations.
The most challenging configurations in I-RAVEN and I-RAVEN-Mesh are \texttt{3x3Grid} and \texttt{Out-InGrid}, in which image panels contain more objects than in the remaining configurations.
Apparently, such setups require stronger reasoning capabilities to correctly identify the rules applied to multiple objects.
Also, the results on the \texttt{Left-Right} and \texttt{Up-Down} configurations are relatively weaker in most regimes.
In these configurations, rules may be applied to both matrix components (left/right and up/down, resp.), increasing the task complexity.
This also concerns the \texttt{Out-InGrid} configuration in the \texttt{A/Size} regime, and the \texttt{Out-InCenter} configuration in the \texttt{A/SizeType} regime.
Results in the \texttt{A/Position} regime are close-to-perfect in configurations comprising a single object in each component (\texttt{Center}, \texttt{Left-Right}, \texttt{Up-Down}, and \texttt{Out-InCenter}) and weaker in the remaining configurations (\texttt{2x2Grid}, \texttt{3x3Grid} and \texttt{Out-InGrid}).
This performance drop can be attributed to the fact that \texttt{Position} attribute can only be effectively applied to the \texttt{2x2Grid}, \texttt{3x3Grid} and \texttt{Out-InGrid} configurations allowing modification of the object's position.
In the remaining configurations its application does not introduce any changes.

\subsection{Progressive knowledge acquisition on I-RAVEN-Mesh}
In the second set of experiments we employ I-RAVEN-Mesh to examine the TL ability of the best performing models.
To this end, we consider variants of partial I-RAVEN-Mesh dataset with a fraction $q \in \{\frac{1}{64},  \ldots, 1\}$ of the training set and compare the performance of a model trained from scratch on a partial dataset to that of a model pre-trained on full I-RAVEN and fine-tuned on the respective part of I-RAVEN-Mesh.
Fig.~\ref{fig:transfer-learning} shows that for $q=\frac{1}{64}$ pre-training RelBase, MRNet, CPCNet, SRAN, and CoPINet on I-RAVEN leads to gains smaller than $15$ p.p., whereas pre-training DRNet, SCL, PredRNet, and STSN improved their accuracy by $50.6$, $34.8$, $23.4$ and $29.6$ p.p., resp.
In addition, TL clearly improved performance of DRNet, SCL, PredRNet, and STSN in all considered settings.
In particular for $q=1$ by $9.0$, $5.9$, $20.5$, and $18.8$ p.p., resp., indicating the models' capacity for knowledge reuse.

\section{Conclusion}
\label{sec:conclusion}
We investigate generalization capabilities of DL models in the AVR domain.
To accelerate research in this area, we propose two RPM benchmarks.
A-I-RAVEN introduces $10$ generalization regimes of variable complexity that assess model's capability to solve matrices with rules applied to novel attributes at various levels of complexity (primary and extended regimes). Contrary to the existing PGM dataset, A-I-RAVEN features compositionality, offers a variety of figure configurations, and above all does not require substantial computational resources.
I-RAVEN-Mesh overlays line-based patterns on top of the RPM, facilitating TL studies.
Experiments on $13$ strong literature AVR models reveal their limitations in terms of generalization.
We believe that the introduced datasets complement existing RPM benchmarks and will foster progress in the AVR area.

\paragraph{Limitations and future work.}
In this work we study generalization and knowledge transfer in contemporary AVR models employing RPM datasets.
While RPMs are by far the most popular AVR tasks, the AVR domain also includes other types of problems not covered in the paper~\cite{malkinski2022review}.
The Machine Number Sense dataset presents visual arithmetic problems~\cite{zhang2020machine}, VAEC defines an extrapolation challenge~\cite{webb2020learning}, while ARC proposes a set of diverse tasks in a few-shot learning setting~\cite{chollet2019measure}.
Similar studies could be performed on problems other than RPMs to test the performance and knowledge transfer abilities of AVR models in other problem settings.

\appendix

\section*{Acknowledgments}
This research was carried out with the support of the Laboratory of Bioinformatics and Computational Genomics and the High Performance Computing Center of the Faculty of Mathematics and Information Science Warsaw University of Technology.
Mikołaj Małkiński was funded by the Warsaw University of Technology within the Excellence Initiative: Research University (IDUB) programme.
This paper builds on the MSc thesis titled "Transfer learning in abstract visual reasoning domain" by Adam Kowalczyk from the Warsaw University of Technology, Warsaw, Poland.

\bibliographystyle{named}
\bibliography{main}

\begin{figure*}
  \centering
  \begin{subfigure}{0.224\textwidth}
    \centering
    \includegraphics[width=\textwidth]{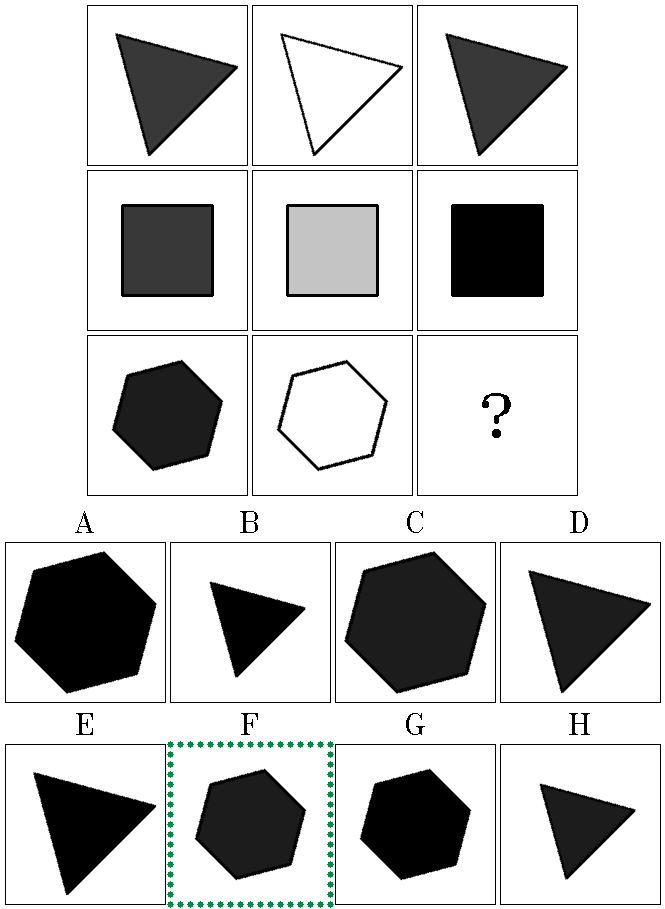}
    \caption{\texttt{A/Type} train}
  \end{subfigure}
  \quad
  \begin{subfigure}{0.224\textwidth}
    \centering
    \includegraphics[width=\textwidth]{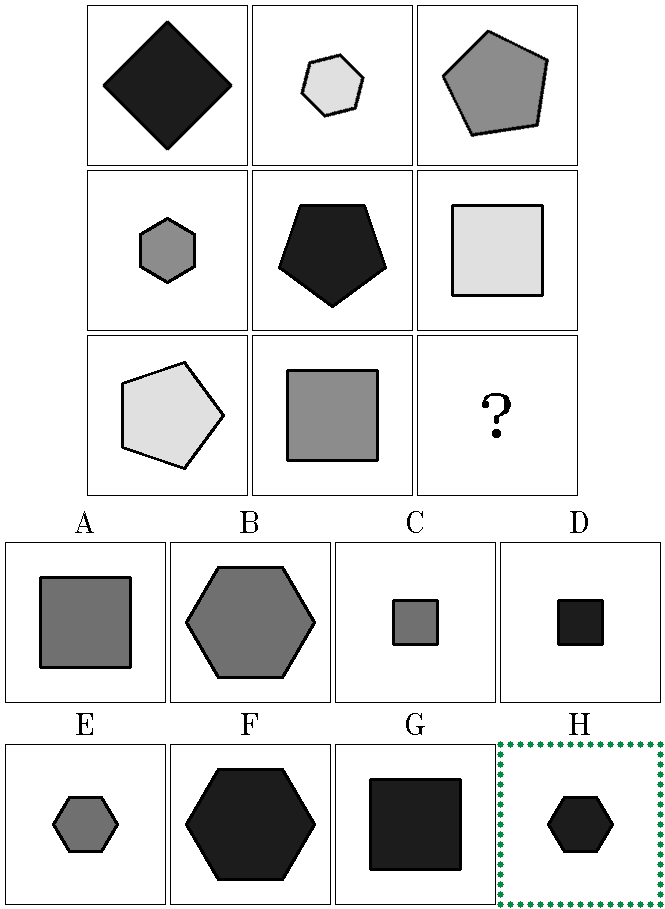}
    \caption{\texttt{A/Type} test}
  \end{subfigure}
  \hfill
  \begin{subfigure}{0.224\textwidth}
    \centering
    \includegraphics[width=\textwidth]{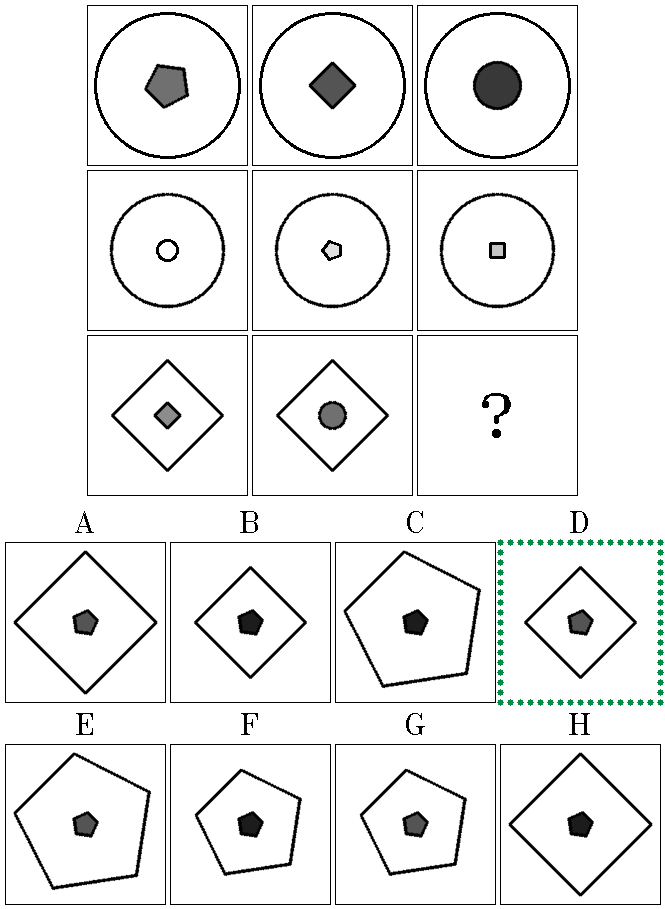}
    \caption{\texttt{A/Size} train}
  \end{subfigure}
  \quad
  \begin{subfigure}{0.224\textwidth}
    \centering
    \includegraphics[width=\textwidth]{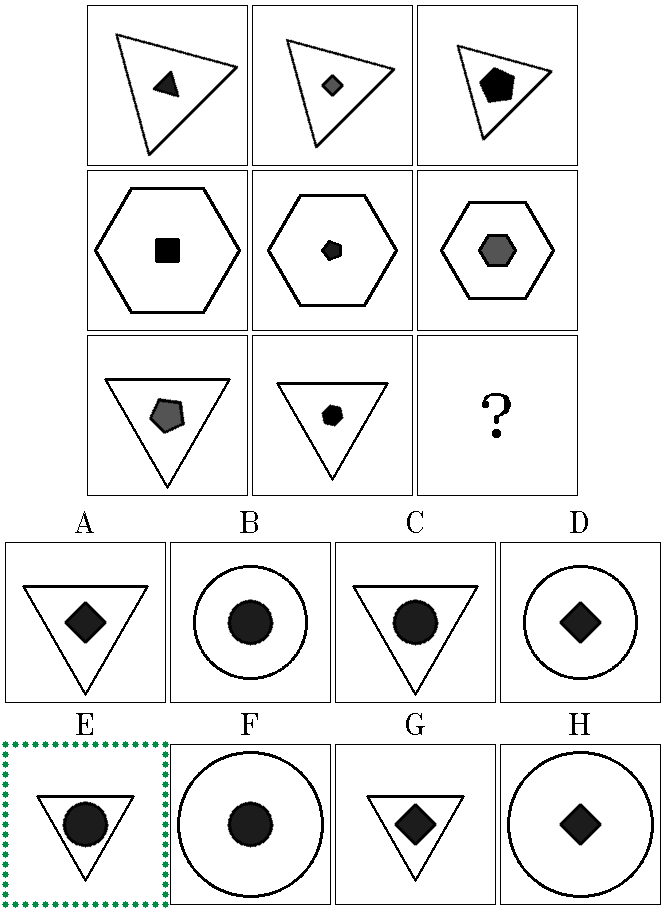}
    \caption{\texttt{A/Size} test}
  \end{subfigure}
  \caption{\textbf{A-I-RAVEN.}
  \underline{Left}: Matrices from the \texttt{A/Type} regime belonging to the \texttt{Center} configuration.
  In (a), object type is constant across rows, while in (b) it's governed by the \texttt{Distribute Three} rule.
  \underline{Right}: Matrices from the \texttt{A/Size} regime belonging to the \texttt{Out-InCenter} configuration.
  In (c), object size is constant across rows in both inner and outer image parts, while in (d) the inner and outer components are governed by the \texttt{Arithmetic} and \texttt{Progression} rules, resp.}
  \label{fig:attributeless-type-size}
\end{figure*}

\begin{figure*}
  \centering
  \begin{subfigure}{0.224\textwidth}
    \centering
    \includegraphics[width=\textwidth]{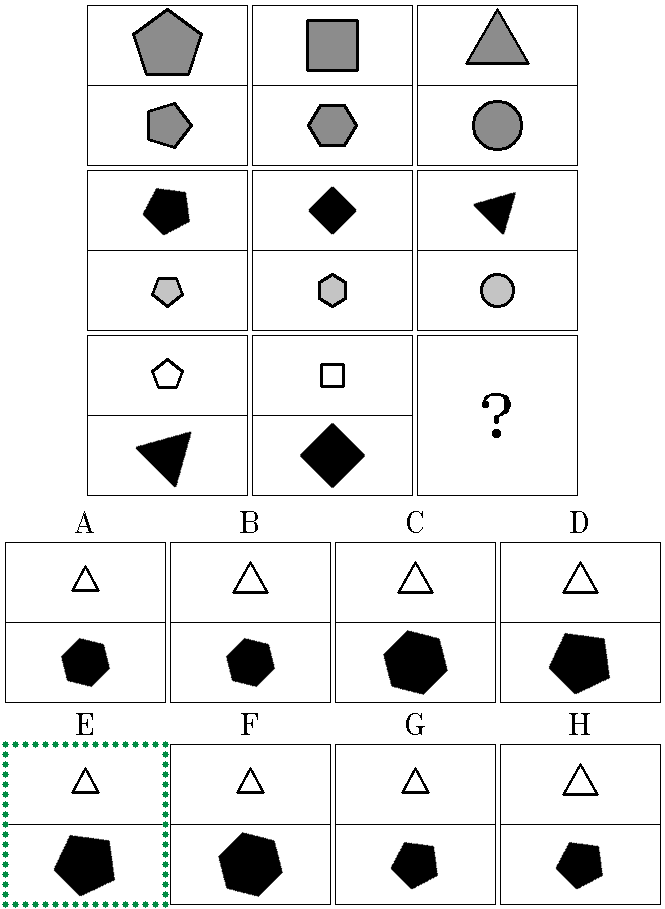}
    \caption{\texttt{A/ColorSize} train}
  \end{subfigure}
  \quad
  \begin{subfigure}{0.224\textwidth}
    \centering
    \includegraphics[width=\textwidth]{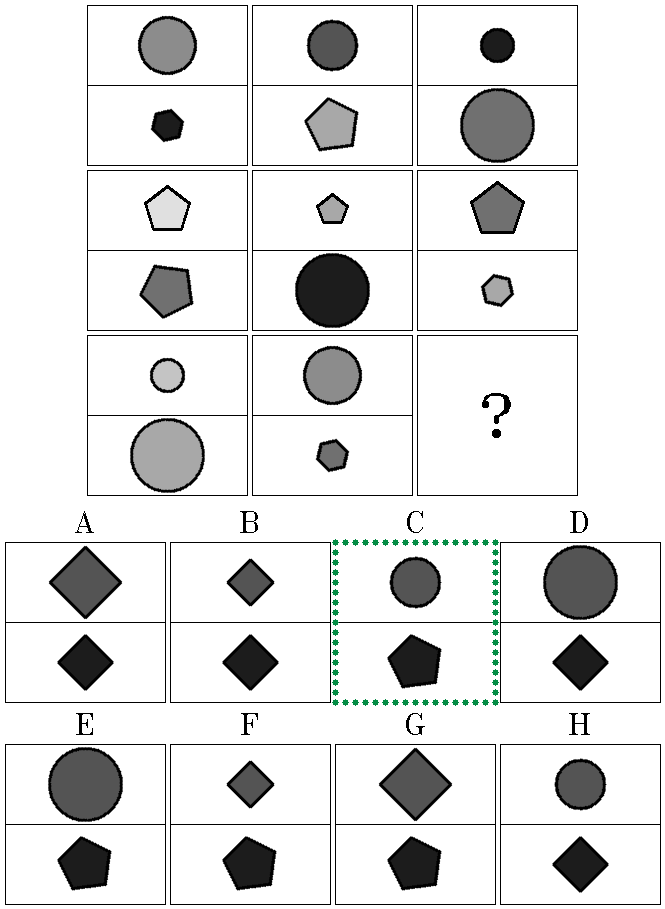}
    \caption{\texttt{A/ColorSize} test}
  \end{subfigure}
  \hfill
  \begin{subfigure}{0.224\textwidth}
    \centering
    \includegraphics[width=\textwidth]{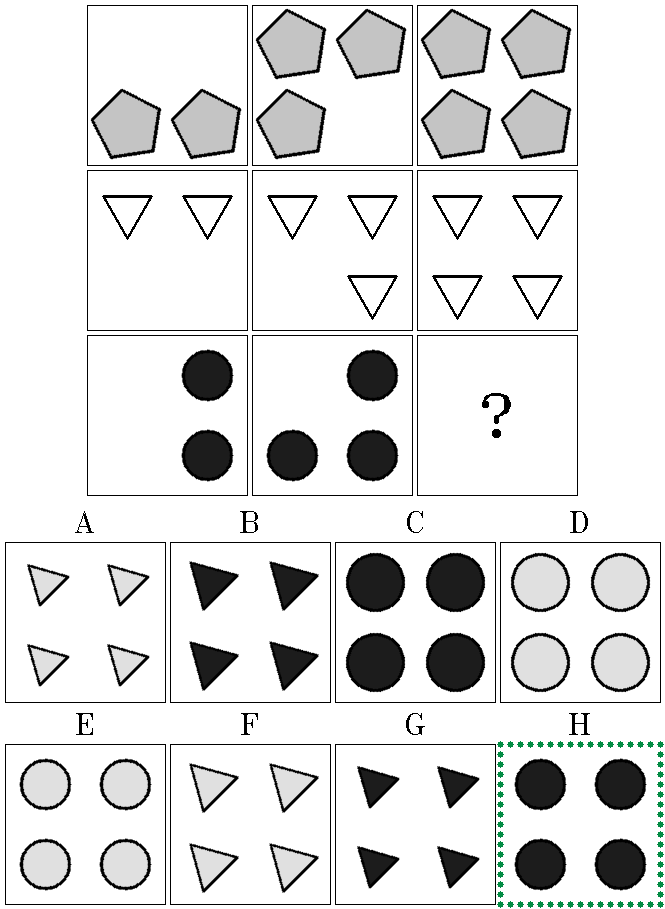}
    \caption{\texttt{A/ColorType} train}
  \end{subfigure}
  \quad
  \begin{subfigure}{0.224\textwidth}
    \centering
    \includegraphics[width=\textwidth]{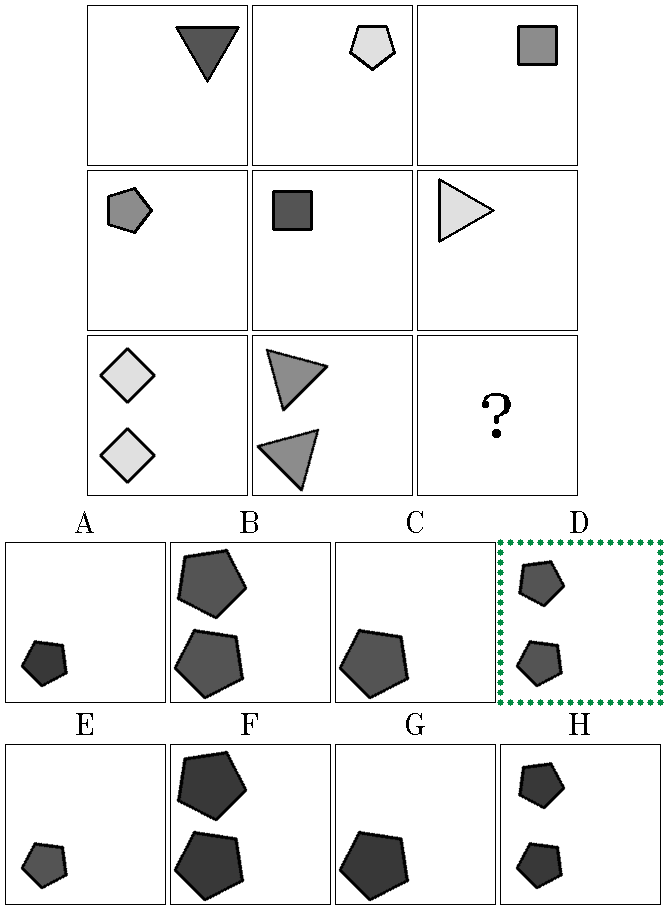}
    \caption{\texttt{A/ColorType} test}
  \end{subfigure}
  \caption{\textbf{A-I-RAVEN.}
  \underline{Left}:
  Matrices from the \texttt{A/ColorSize} regime belonging to the \texttt{Up-Down} configuration.
  In (a), object color and size is constant across rows in both components, while in (b) they are governed by \texttt{Progression} and \texttt{Distribute Three} in the upper component, resp., and by \texttt{Distribute Three} in the lower one.
  \underline{Right}:
  Matrices from the \texttt{A/ColorType} regime belonging to the \texttt{2x2 Grid} configuration.
  In (c), object color and type is constant across rows, while in (d) they are governed by the \texttt{Distribute Three} rule.
  }
  \label{fig:attributeless-color-size-and-color-type}
\end{figure*}

\begin{figure*}
  \centering
  \begin{subfigure}{0.224\textwidth}
    \centering
    \includegraphics[width=\textwidth]{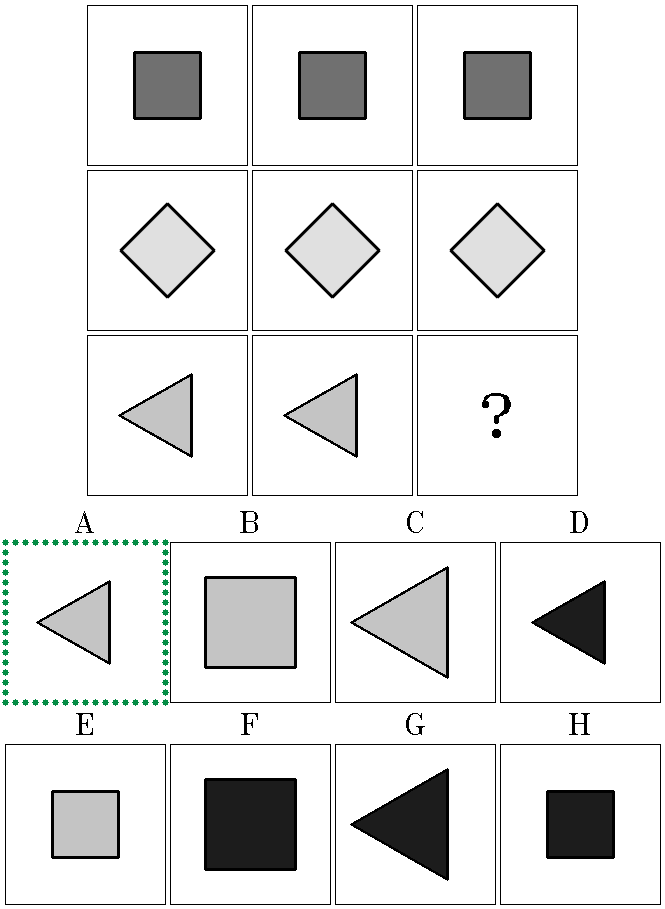}
    \caption{\texttt{A/SizeType} train}
  \end{subfigure}
  \quad
  \begin{subfigure}{0.224\textwidth}
    \centering
    \includegraphics[width=\textwidth]{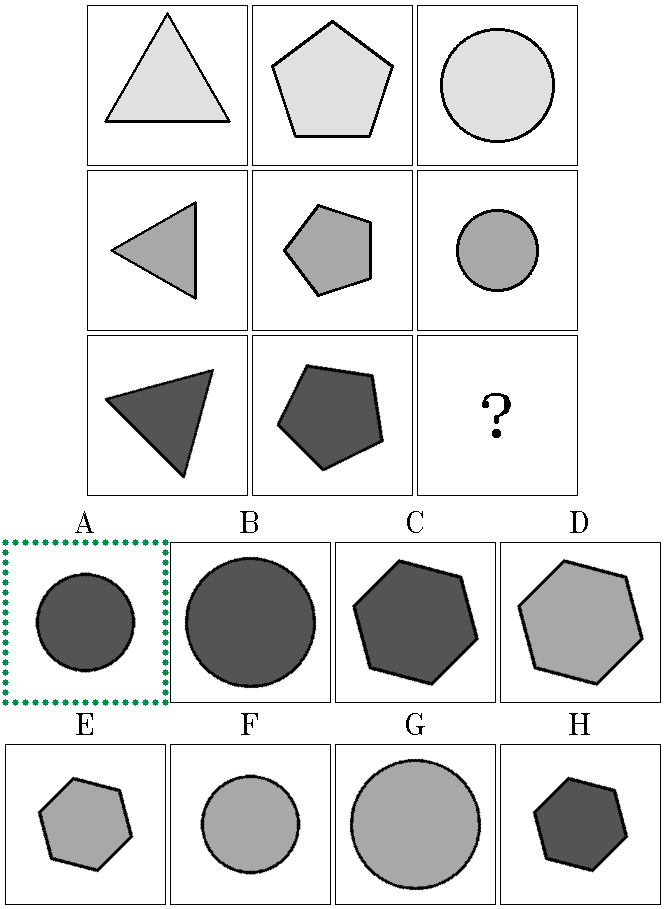}
    \caption{\texttt{A/SizeType} test}
  \end{subfigure}
  \hfill
  \begin{subfigure}{0.224\textwidth}
    \centering
    \includegraphics[width=\textwidth]{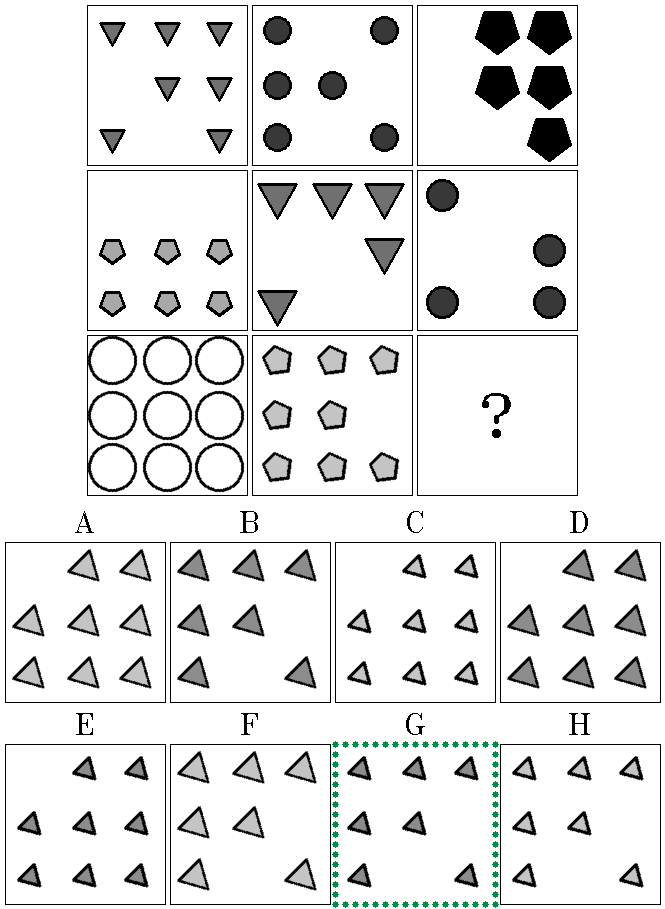}
    \caption{\texttt{A/Color-P} train}
  \end{subfigure}
  \quad
  \begin{subfigure}{0.224\textwidth}
    \centering
    \includegraphics[width=\textwidth]{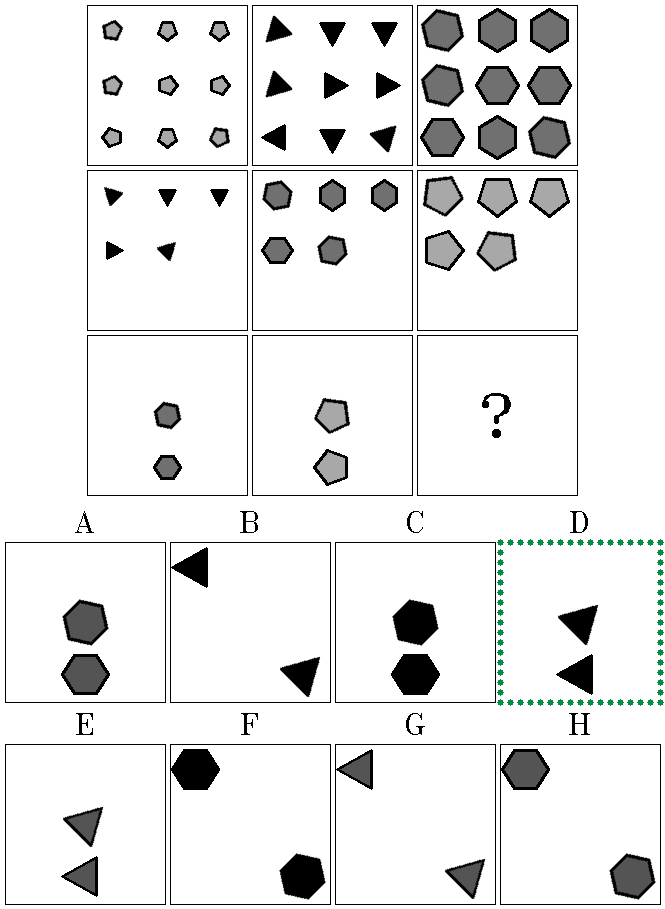}
    \caption{\texttt{A/Color-P} test}
  \end{subfigure}
  \caption{\textbf{A-I-RAVEN.}
  \underline{Left}:
  Matrices from the \texttt{A/SizeType} regime belonging to the \texttt{Center} configuration.
  In (a), object size and type are constant across rows, while in (b) they are governed by the \texttt{Progression} rule.
  \underline{Right}:
  Matrices from the \texttt{A/Color-Progression} regime belonging to the \texttt{3x3 Grid} configuration.
  In (c), object color is governed by the \texttt{Progression} rule, while in (d) by \texttt{Distribute Three}.
  }
  \label{fig:attributeless-size-type-and-color-progression}
\end{figure*}

\begin{figure*}
  \centering
  \begin{subfigure}{0.224\textwidth}
    \centering
    \includegraphics[width=\textwidth]{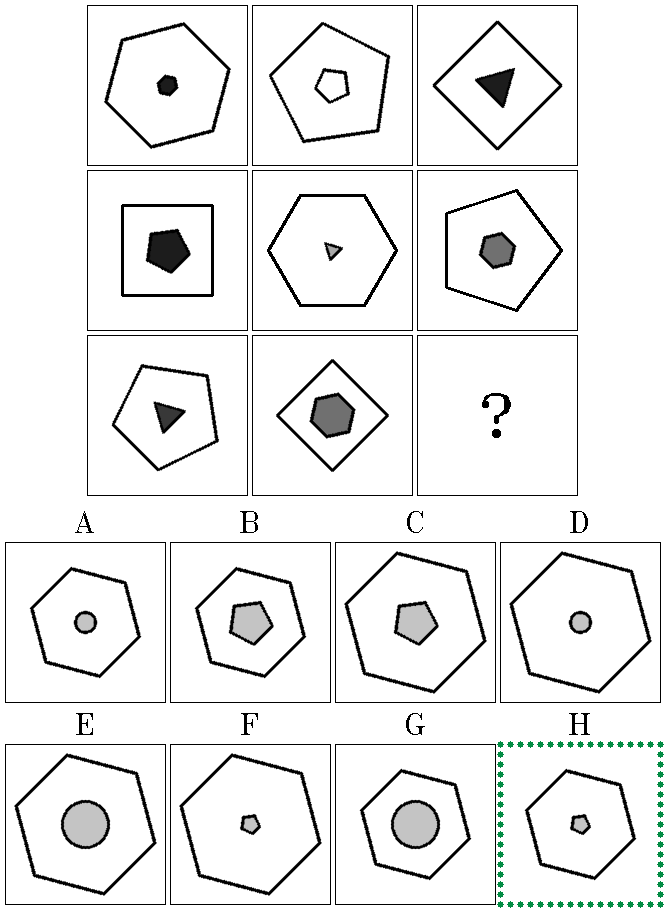}
    \caption{\texttt{A/Color-A} train}
  \end{subfigure}
  \quad
  \begin{subfigure}{0.224\textwidth}
    \centering
    \includegraphics[width=\textwidth]{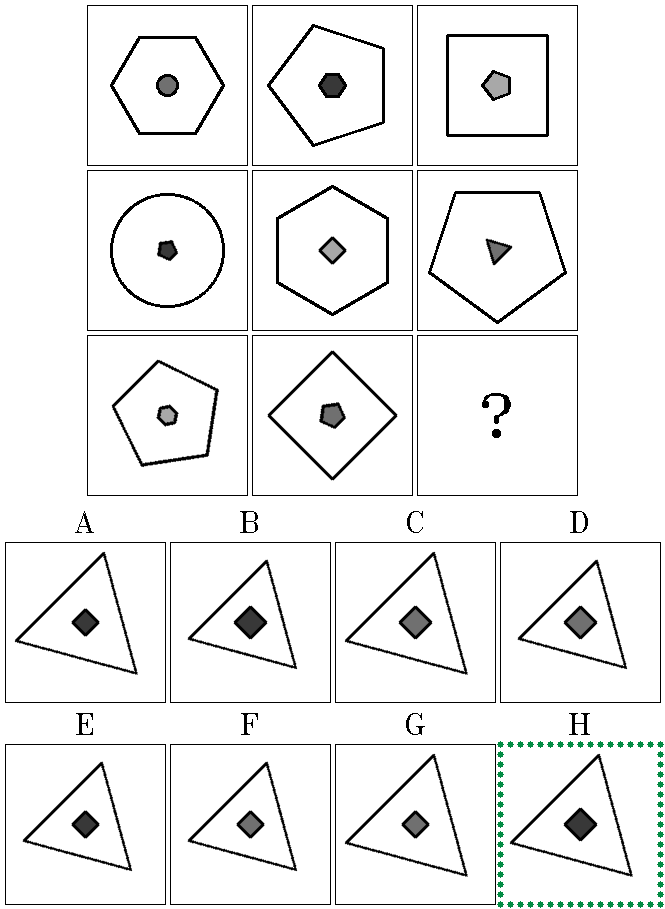}
    \caption{\texttt{A/Color-A} test}
  \end{subfigure}
  \hfill
  \begin{subfigure}{0.224\textwidth}
    \centering
    \includegraphics[width=\textwidth]{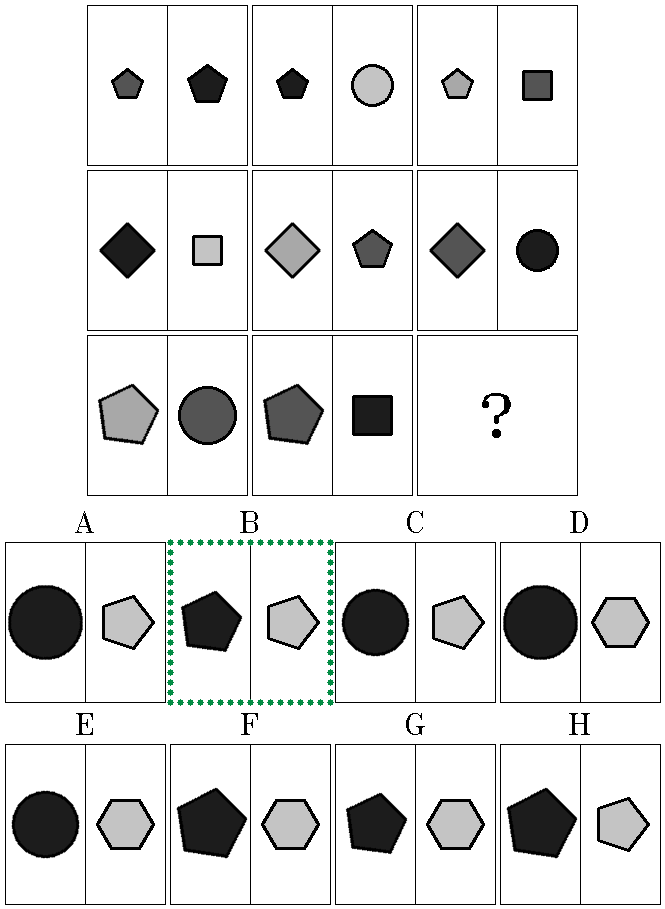}
    \caption{\texttt{A/Color-D3} train}
  \end{subfigure}
  \quad
  \begin{subfigure}{0.224\textwidth}
    \centering
    \includegraphics[width=\textwidth]{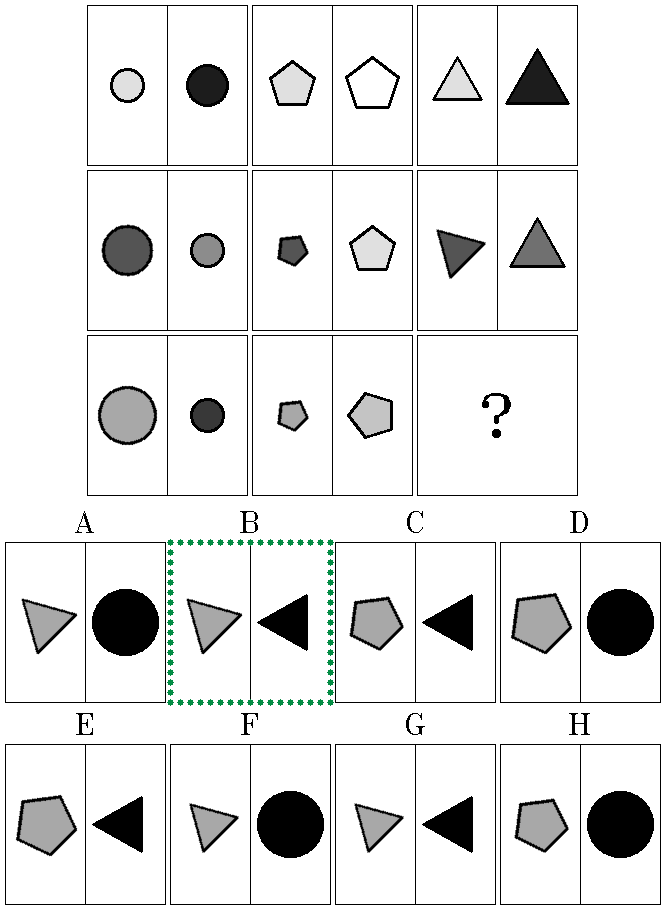}
    \caption{\texttt{A/Color-D3} test}
  \end{subfigure}
  \caption{\textbf{A-I-RAVEN.}
  \underline{Left}:
  Matrices from the \texttt{A/Color-Arithmetic} regime belonging to the \texttt{Out-InCenter} configuration.
  In (a), object color in the inner component is governed by \texttt{Arithmetic}, while in (b) it is governed by \texttt{Distribute Three}.
  \underline{Right}:
  Matrices from the \texttt{A/Color-DistributeThree} regime belonging to the \texttt{Left-Right} configuration.
  In (c), object color is governed by the \texttt{Distribute Three} rule in both components, while in (d) by \texttt{Constant} in the left component and by \texttt{Arithmetic} in the right one.
  }
  \label{fig:attributeless-color-arithmetic-and-color-distributethree}
\end{figure*}

\begin{figure*}
  \centering
  \begin{subfigure}{0.224\textwidth}
    \centering
    \includegraphics[width=\textwidth]{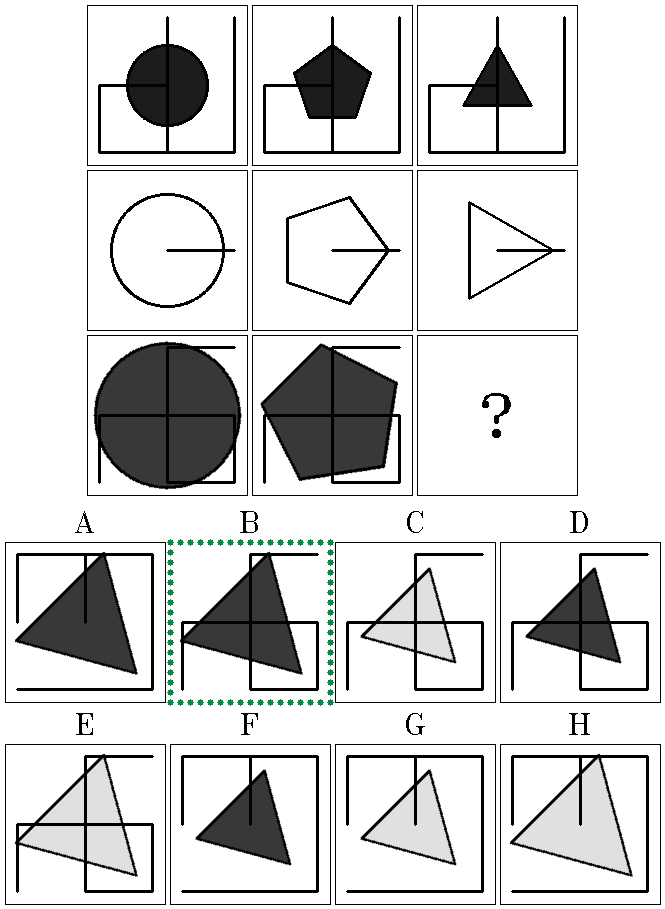}
    \caption{\texttt{Constant}}
  \end{subfigure}
  \hfill
  \begin{subfigure}{0.224\textwidth}
    \centering
    \includegraphics[width=\textwidth]{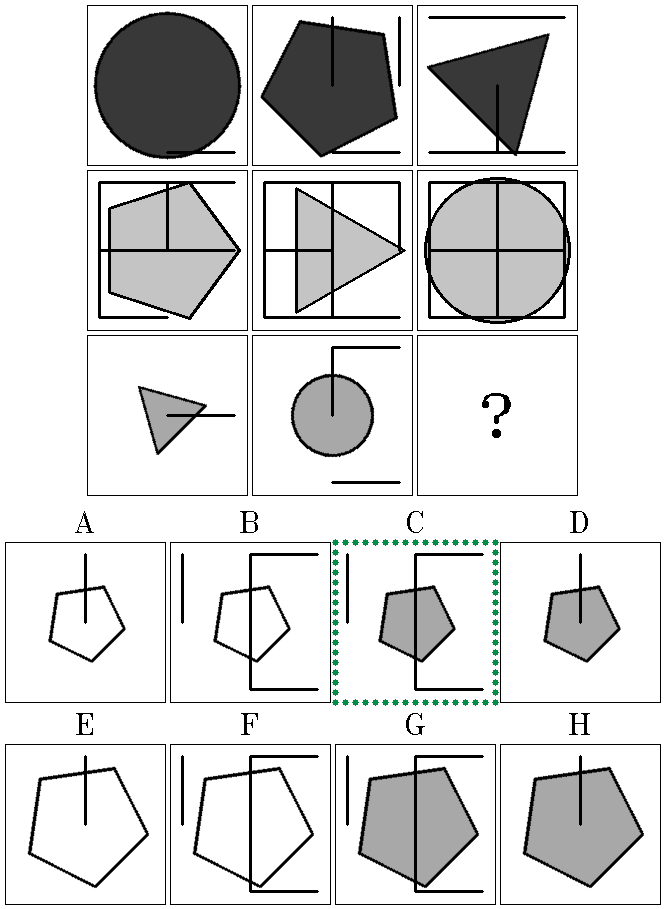}
    \caption{\texttt{Progression}}
  \end{subfigure}
  \hfill
  \begin{subfigure}{0.224\textwidth}
    \centering
    \includegraphics[width=\textwidth]{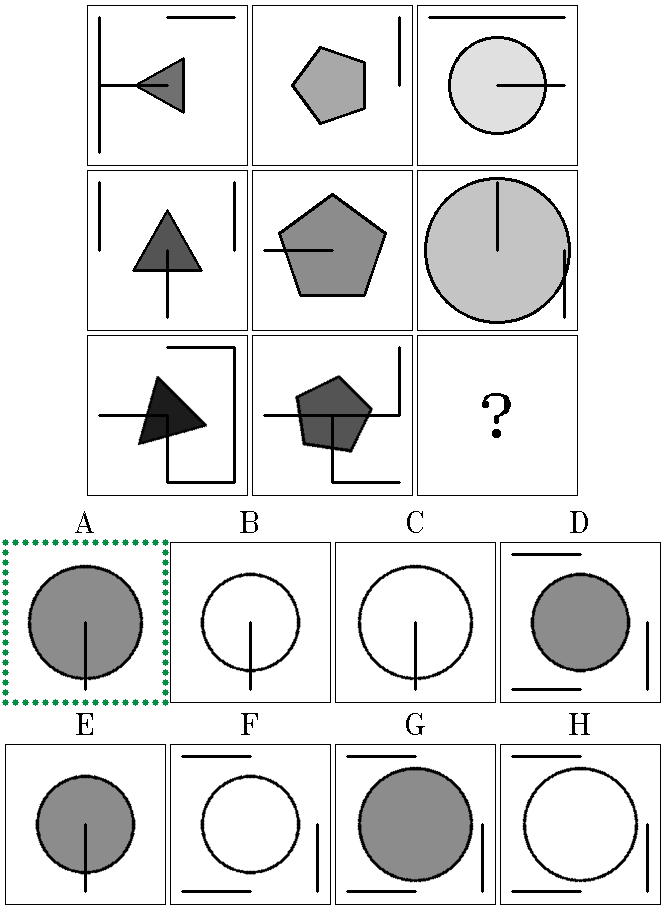}
    \caption{\texttt{Arithmetic}}
  \end{subfigure}
  \hfill
  \begin{subfigure}{0.224\textwidth}
    \centering
    \includegraphics[width=\textwidth]{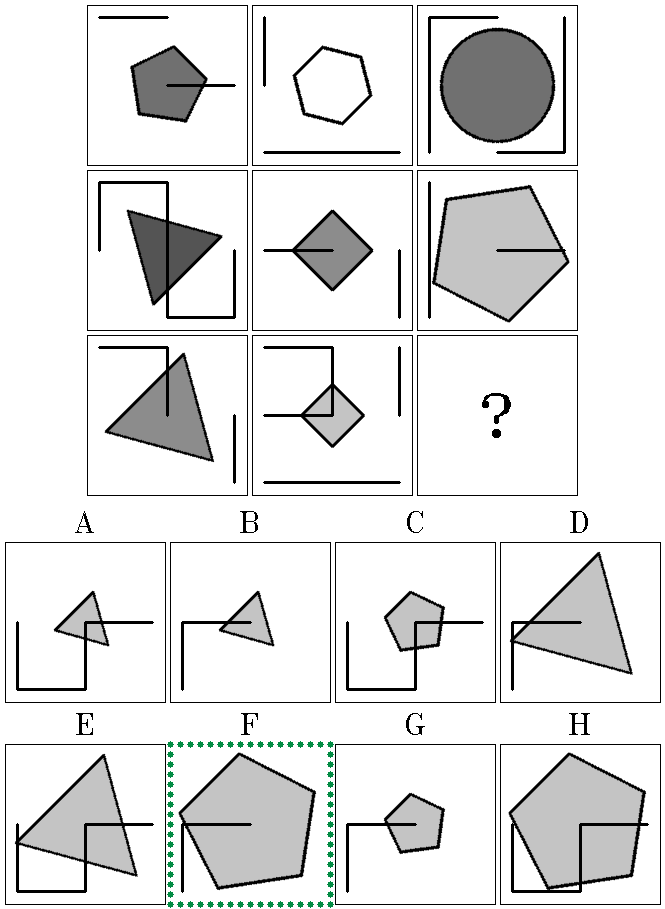}
    \caption{\texttt{Distribute Three}}
  \end{subfigure}
  \caption{\textbf{I-RAVEN-Mesh.}
  The examples showcase matrices with the \texttt{Number} attribute of the mesh component governed by all applicable rules.
  (a) Line number is constant in each row.
  (b) The number of lines increases by $2$ from left to right.
  (c) The number of lines in the third column is the difference between the number of lines in the second and first columns.
  (d) The numbers of lines in each row compose a set $\{2, 3, 6\}$.}
  \label{fig:mesh-number}
\end{figure*}

\section{Additional matrix examples}
\label{sec:examples}

Figure~\ref{fig:attributeless-type-size} presents matrix examples from $\texttt{A/Type}$ and $\texttt{A/Size}$, the primary regimes of A-I-RAVEN. Figures~\ref{fig:attributeless-color-size-and-color-type},~\ref{fig:attributeless-size-type-and-color-progression} and~\ref{fig:attributeless-color-arithmetic-and-color-distributethree}  depict matrix examples from the extended regimes of A-I-RAVEN:
$\texttt{A/ColorSize}$ and $\texttt{A/ColorType}$ (Fig.~\ref{fig:attributeless-color-size-and-color-type}),
$\texttt{A/SizeType}$ and $\texttt{A/Color-Progression}$ (Fig.~\ref{fig:attributeless-size-type-and-color-progression}), and 
$\texttt{A/Color-Arithmetic}$ and $\texttt{A/Color-DistributeThree}$ (Fig.~\ref{fig:attributeless-color-arithmetic-and-color-distributethree}).
Figure~\ref{fig:mesh-number} presents matrix examples from I-RAVEN-Mesh concerning the $\texttt{Number}$ attribute.



\section{Auxiliary training}
\label{sec:dataset-details}
As discussed in Section~\ref{sec:experiments}, we use sparse encoding~\cite{malkinski2020multi} to represent the set of matrix rules as a vector $r \in \mathbb{R}^{d_r}$, such that $d_r = 48$ for I-RAVEN-Mesh and $d_r = 40$ otherwise.
The set of rules $\mathcal{R}$ in I-RAVEN is \{\texttt{Constant}, \texttt{Progression}, \texttt{Arithmetic}, \texttt{Distribute Three}\} and the set of attributes $\mathcal{A}$ is \{\texttt{Position}, \texttt{Number}, \texttt{Type}, \texttt{Size}, \texttt{Color}\}.
It follows that there is $\vert \mathcal{R} \vert \times \vert \mathcal{A} \vert = 20$ unique rule--attribute pairs.
In addition, the \texttt{Left-Right}, \texttt{Up-Down}, \texttt{Out-InCenter}, and \texttt{Out-InGrid} configurations in I-RAVEN comprise two components in which rules exist independently, e.g., the \texttt{Left-Right} component contains matrices with separate rules applied to the left and right sides.
This gives an upper bound of $40$ rule--attribute combinations in each configuration.

As discussed in Section~\ref{sec:i-raven-mesh} and presented in Table~\ref{tab:mesh-description}, the Mesh component introduced in I-RAVEN-Mesh comprises two attributes and four rules, leading to a total of $48$ rule--attribute combinations per configuration.
As an example, in the \texttt{Up-Down} configuration of I-RAVEN-Mesh, there are $20$ rule--attribute combinations for the upper component, another $20$ for the lower component, and $8$ for the Mesh component.

The sparse encoding encodes each rule in a matrix as a one-hot vector and applies the OR operation to the set of one-hot vectors, producing a multi-hot representation of matrix rules.

\section{Extended results}
\label{sec:extended-results}
Tables~\ref{tab:stl-aux-i-raven} --~\ref{tab:stl-aux-color-distributethree} present the results (mean and standard deviation) of all considered models on test and validation splits and the difference between these two splits for particular datasets/regimes.

Tables~\ref{tab:stl-configurations-i-raven} --~\ref{tab:stl-configurations-color-distributethree} present 
the results (mean and standard deviation) of all considered models in detail for all matrix configurations.

\FloatBarrier


\begin{table}[t]
\centering\small
\begin{tabular}{l|ccc}
\toprule
 & Test & Val & Test $-$ Val \\
\midrule
ALANS & $27.0$ \scriptsize$(\pm\,8.4)$ & $27.0$ \scriptsize$(\pm\,8.6)$ & $+\ \ 0.1$ \\
CPCNet & $70.4$ \scriptsize$(\pm\,6.4)$ & $69.6$ \scriptsize$(\pm\,6.9)$ & $+\ \ 0.7$ \\
CNN-LSTM & $27.5$ \scriptsize$(\pm\,1.5)$ & $27.4$ \scriptsize$(\pm\,1.7)$ & $+\ \ 0.1$ \\
CoPINet & $43.2$ \scriptsize$(\pm\,0.1)$ & $42.5$ \scriptsize$(\pm\,0.6)$ & $+\ \ 0.7$ \\
DRNet & $\textbf{90.9}$ \scriptsize$(\pm\,1.1)$ & $\textbf{90.8}$ \scriptsize$(\pm\,1.2)$ & $+\ \ 0.2$ \\
MRNet & $86.7$ \scriptsize$(\pm\,2.3)$ & $86.3$ \scriptsize$(\pm\,2.0)$ & $+\ \ 0.5$ \\
PrAE & $19.5$ \scriptsize$(\pm\,0.4)$ & $19.4$ \scriptsize$(\pm\,0.8)$ & $+\ \ 0.0$ \\
PredRNet & $88.8$ \scriptsize$(\pm\,1.8)$ & $88.3$ \scriptsize$(\pm\,1.9)$ & $+\ \ 0.5$ \\
RelBase & $\underline{89.6}$ \scriptsize$(\pm\,0.6)$ & $\underline{89.5}$ \scriptsize$(\pm\,0.5)$ & $+\ \ 0.1$ \\
SCL & $83.4$ \scriptsize$(\pm\,2.5)$ & $83.0$ \scriptsize$(\pm\,2.5)$ & $+\ \ 0.4$ \\
SRAN & $58.2$ \scriptsize$(\pm\,1.6)$ & $58.0$ \scriptsize$(\pm\,1.3)$ & $+\ \ 0.2$ \\
STSN & $59.0$ \scriptsize$(\pm\,18.5)$ & $59.1$ \scriptsize$(\pm\,18.4)$ & $-\ \ 0.1$ \\
WReN & $18.4$ \scriptsize$(\pm\,0.0)$ & $18.5$ \scriptsize$(\pm\,0.3)$ & $-\ \ 0.1$ \\
\bottomrule
\end{tabular}
\caption{\textbf{I-RAVEN.}}
\label{tab:stl-aux-i-raven}
\end{table}

\begin{table}[t]
\centering\small
\begin{tabular}{l|ccc}
\toprule
 & Test & Val & Test $-$ Val \\
\midrule
ALANS & $15.9$ \scriptsize$(\pm\,2.6)$ & $17.1$ \scriptsize$(\pm\,3.6)$ & $-\ \ 1.3$ \\
CPCNet & $66.6$ \scriptsize$(\pm\,5.1)$ & $66.5$ \scriptsize$(\pm\,5.4)$ & $+\ \ 0.1$ \\
CNN-LSTM & $28.9$ \scriptsize$(\pm\,0.4)$ & $29.3$ \scriptsize$(\pm\,0.6)$ & $-\ \ 0.4$ \\
CoPINet & $41.1$ \scriptsize$(\pm\,0.3)$ & $41.3$ \scriptsize$(\pm\,0.2)$ & $-\ \ 0.2$ \\
DRNet & $\underline{83.9}$ \scriptsize$(\pm\,2.7)$ & $\underline{84.2}$ \scriptsize$(\pm\,2.6)$ & $-\ \ 0.3$ \\
MRNet & $79.5$ \scriptsize$(\pm\,2.0)$ & $80.5$ \scriptsize$(\pm\,2.5)$ & $-\ \ 1.0$ \\
PrAE & $33.2$ \scriptsize$(\pm\,0.4)$ & $33.0$ \scriptsize$(\pm\,0.9)$ & $+\ \ 0.1$ \\
PredRNet & $59.2$ \scriptsize$(\pm\,6.4)$ & $59.3$ \scriptsize$(\pm\,6.9)$ & $-\ \ 0.0$ \\
RelBase & $\textbf{84.9}$ \scriptsize$(\pm\,4.4)$ & $\textbf{85.0}$ \scriptsize$(\pm\,4.5)$ & $-\ \ 0.1$ \\
SCL & $80.9$ \scriptsize$(\pm\,1.5)$ & $81.0$ \scriptsize$(\pm\,1.5)$ & $-\ \ 0.1$ \\
SRAN & $57.8$ \scriptsize$(\pm\,0.2)$ & $58.0$ \scriptsize$(\pm\,0.3)$ & $-\ \ 0.2$ \\
STSN & $48.7$ \scriptsize$(\pm\,11.5)$ & $48.8$ \scriptsize$(\pm\,10.9)$ & $-\ \ 0.1$ \\
WReN & $25.7$ \scriptsize$(\pm\,0.2)$ & $25.6$ \scriptsize$(\pm\,0.4)$ & $+\ \ 0.0$ \\
\bottomrule
\end{tabular}
\caption{\textbf{I-RAVEN-Mesh.}}
\label{tab:stl-aux-mesh}
\end{table}

\begin{table}[t]
\centering\small
\begin{tabular}{l|ccc}
\toprule
 & Test & Val & Test $-$ Val \\
\midrule
ALANS & $15.2$ \scriptsize$(\pm\,1.4)$ & $16.4$ \scriptsize$(\pm\,2.1)$ & $-\ \ 1.2$ \\
CPCNet & $51.2$ \scriptsize$(\pm\,3.8)$ & $77.0$ \scriptsize$(\pm\,6.1)$ & $-25.7$ \\
CNN-LSTM & $17.0$ \scriptsize$(\pm\,3.1)$ & $31.0$ \scriptsize$(\pm\,4.4)$ & $-13.9$ \\
CoPINet & $32.5$ \scriptsize$(\pm\,0.2)$ & $49.9$ \scriptsize$(\pm\,0.7)$ & $-17.4$ \\
DRNet & $\textbf{70.0}$ \scriptsize$(\pm\,1.6)$ & $\textbf{95.4}$ \scriptsize$(\pm\,0.2)$ & $-25.4$ \\
MRNet & $33.6$ \scriptsize$(\pm\,8.2)$ & $86.2$ \scriptsize$(\pm\,6.6)$ & $-52.6$ \\
PrAE & $47.9$ \scriptsize$(\pm\,0.9)$ & $60.9$ \scriptsize$(\pm\,1.4)$ & $-13.0$ \\
PredRNet & $59.4$ \scriptsize$(\pm\,1.0)$ & $92.2$ \scriptsize$(\pm\,1.0)$ & $-32.9$ \\
RelBase & $\underline{67.4}$ \scriptsize$(\pm\,2.7)$ & $\underline{95.2}$ \scriptsize$(\pm\,0.4)$ & $-27.8$ \\
SCL & $65.1$ \scriptsize$(\pm\,2.0)$ & $84.4$ \scriptsize$(\pm\,0.5)$ & $-19.2$ \\
SRAN & $38.3$ \scriptsize$(\pm\,1.0)$ & $63.7$ \scriptsize$(\pm\,0.3)$ & $-25.4$ \\
STSN & $39.3$ \scriptsize$(\pm\,6.9)$ & $71.3$ \scriptsize$(\pm\,17.0)$ & $-32.0$ \\
WReN & $16.9$ \scriptsize$(\pm\,0.5)$ & $23.2$ \scriptsize$(\pm\,0.8)$ & $-\ \ 6.3$ \\
\bottomrule
\end{tabular}
\caption{\textbf{\texttt{A/Color}.}}
\label{tab:stl-aux-color}
\end{table}

\begin{table}[t]
\centering\small
\begin{tabular}{l|ccc}
\toprule
 & Test & Val & Test $-$ Val \\
\midrule
ALANS & $16.0$ \scriptsize$(\pm\,1.0)$ & $15.2$ \scriptsize$(\pm\,1.3)$ & $+\ \ 0.8$ \\
CPCNet & $68.3$ \scriptsize$(\pm\,4.0)$ & $90.6$ \scriptsize$(\pm\,5.3)$ & $-22.3$ \\
CNN-LSTM & $24.0$ \scriptsize$(\pm\,2.9)$ & $36.4$ \scriptsize$(\pm\,3.7)$ & $-12.4$ \\
CoPINet & $41.3$ \scriptsize$(\pm\,1.6)$ & $54.7$ \scriptsize$(\pm\,1.7)$ & $-13.4$ \\
DRNet & $\textbf{77.5}$ \scriptsize$(\pm\,0.9)$ & $\textbf{97.8}$ \scriptsize$(\pm\,0.1)$ & $-20.2$ \\
MRNet & $62.6$ \scriptsize$(\pm\,2.6)$ & $94.4$ \scriptsize$(\pm\,7.0)$ & $-31.8$ \\
PrAE & $68.2$ \scriptsize$(\pm\,3.3)$ & $80.1$ \scriptsize$(\pm\,3.4)$ & $-12.0$ \\
PredRNet & $73.7$ \scriptsize$(\pm\,0.7)$ & $\underline{97.4}$ \scriptsize$(\pm\,0.5)$ & $-23.6$ \\
RelBase & $76.6$ \scriptsize$(\pm\,0.3)$ & $97.0$ \scriptsize$(\pm\,0.2)$ & $-20.4$ \\
SCL & $\underline{76.7}$ \scriptsize$(\pm\,7.1)$ & $94.7$ \scriptsize$(\pm\,5.3)$ & $-18.0$ \\
SRAN & $56.9$ \scriptsize$(\pm\,0.7)$ & $75.6$ \scriptsize$(\pm\,1.4)$ & $-18.8$ \\
STSN & $36.1$ \scriptsize$(\pm\,19.9)$ & $50.7$ \scriptsize$(\pm\,27.3)$ & $-14.6$ \\
WReN & $17.3$ \scriptsize$(\pm\,0.4)$ & $23.3$ \scriptsize$(\pm\,0.5)$ & $-\ \ 6.0$ \\
\bottomrule
\end{tabular}
\caption{\textbf{\texttt{A/Position}.}}
\label{tab:stl-aux-position}
\end{table}

\begin{table}[t]
\centering\small
\begin{tabular}{l|ccc}
\toprule
 & Test & Val & Test $-$ Val \\
\midrule
ALANS & $23.3$ \scriptsize$(\pm\,6.5)$ & $24.6$ \scriptsize$(\pm\,10.6)$ & $-\ \ 1.2$ \\
CPCNet & $43.5$ \scriptsize$(\pm\,3.5)$ & $79.2$ \scriptsize$(\pm\,2.1)$ & $-35.6$ \\
CNN-LSTM & $13.6$ \scriptsize$(\pm\,1.4)$ & $37.2$ \scriptsize$(\pm\,3.1)$ & $-23.6$ \\
CoPINet & $21.8$ \scriptsize$(\pm\,0.2)$ & $60.7$ \scriptsize$(\pm\,0.1)$ & $-38.9$ \\
DRNet & $\underline{54.3}$ \scriptsize$(\pm\,3.0)$ & $\underline{93.1}$ \scriptsize$(\pm\,0.5)$ & $-38.8$ \\
MRNet & $20.6$ \scriptsize$(\pm\,5.0)$ & $88.1$ \scriptsize$(\pm\,2.5)$ & $-67.5$ \\
PrAE & $41.3$ \scriptsize$(\pm\,1.8)$ & $59.7$ \scriptsize$(\pm\,1.8)$ & $-18.4$ \\
PredRNet & $47.5$ \scriptsize$(\pm\,1.3)$ & $92.5$ \scriptsize$(\pm\,0.8)$ & $-45.0$ \\
RelBase & $51.1$ \scriptsize$(\pm\,2.4)$ & $92.9$ \scriptsize$(\pm\,0.4)$ & $-41.8$ \\
SCL & $\textbf{65.6}$ \scriptsize$(\pm\,2.4)$ & $\textbf{94.0}$ \scriptsize$(\pm\,2.6)$ & $-28.4$ \\
SRAN & $34.4$ \scriptsize$(\pm\,3.0)$ & $78.1$ \scriptsize$(\pm\,0.2)$ & $-43.7$ \\
STSN & $38.4$ \scriptsize$(\pm\,16.6)$ & $68.9$ \scriptsize$(\pm\,34.1)$ & $-30.5$ \\
WReN & $12.4$ \scriptsize$(\pm\,0.5)$ & $29.5$ \scriptsize$(\pm\,0.5)$ & $-17.1$ \\
\bottomrule
\end{tabular}
\caption{\textbf{\texttt{A/Size}.}}
\label{tab:stl-aux-size}
\end{table}

\begin{table}[t]
\centering\small
\begin{tabular}{l|ccc}
\toprule
 & Test & Val & Test $-$ Val \\
\midrule
ALANS & $19.0$ \scriptsize$(\pm\,3.4)$ & $25.2$ \scriptsize$(\pm\,5.2)$ & $-\ \ 6.3$ \\
CPCNet & $38.6$ \scriptsize$(\pm\,4.3)$ & $85.2$ \scriptsize$(\pm\,3.3)$ & $-46.7$ \\
CNN-LSTM & $14.5$ \scriptsize$(\pm\,0.8)$ & $38.5$ \scriptsize$(\pm\,0.8)$ & $-24.0$ \\
CoPINet & $19.8$ \scriptsize$(\pm\,0.9)$ & $58.8$ \scriptsize$(\pm\,2.6)$ & $-39.0$ \\
DRNet & $\underline{44.3}$ \scriptsize$(\pm\,0.8)$ & $\underline{93.9}$ \scriptsize$(\pm\,0.3)$ & $-49.6$ \\
MRNet & $19.4$ \scriptsize$(\pm\,0.3)$ & $93.0$ \scriptsize$(\pm\,0.7)$ & $-73.7$ \\
PrAE & $37.0$ \scriptsize$(\pm\,1.7)$ & $61.1$ \scriptsize$(\pm\,1.4)$ & $-24.1$ \\
PredRNet & $40.2$ \scriptsize$(\pm\,1.3)$ & $93.9$ \scriptsize$(\pm\,0.1)$ & $-53.7$ \\
RelBase & $44.1$ \scriptsize$(\pm\,1.0)$ & $93.4$ \scriptsize$(\pm\,0.2)$ & $-49.3$ \\
SCL & $\textbf{49.5}$ \scriptsize$(\pm\,1.8)$ & $\textbf{95.9}$ \scriptsize$(\pm\,0.7)$ & $-46.4$ \\
SRAN & $30.7$ \scriptsize$(\pm\,2.2)$ & $78.8$ \scriptsize$(\pm\,0.7)$ & $-48.1$ \\
STSN & $39.1$ \scriptsize$(\pm\,5.0)$ & $66.2$ \scriptsize$(\pm\,17.7)$ & $-27.0$ \\
WReN & $15.1$ \scriptsize$(\pm\,0.7)$ & $23.4$ \scriptsize$(\pm\,0.6)$ & $-\ \ 8.2$ \\
\bottomrule
\end{tabular}
\caption{\textbf{\texttt{A/Type}.}}
\label{tab:stl-aux-type}
\end{table}

\begin{table}[t]
\centering\small
\begin{tabular}{l|ccc}
\toprule
 & Test & Val & Test $-$ Val \\
\midrule
ALANS & $15.1$ \scriptsize$(\pm\,3.3)$ & $16.4$ \scriptsize$(\pm\,4.3)$ & $-\ \ 1.3$ \\
CPCNet & $33.0$ \scriptsize$(\pm\,5.3)$ & $86.1$ \scriptsize$(\pm\,1.4)$ & $-53.1$ \\
CNN-LSTM & $13.4$ \scriptsize$(\pm\,0.9)$ & $53.1$ \scriptsize$(\pm\,6.1)$ & $-39.8$ \\
CoPINet & $18.3$ \scriptsize$(\pm\,0.3)$ & $71.7$ \scriptsize$(\pm\,0.3)$ & $-53.4$ \\
DRNet & $\underline{38.3}$ \scriptsize$(\pm\,0.5)$ & $\textbf{95.9}$ \scriptsize$(\pm\,0.6)$ & $-57.6$ \\
MRNet & $18.7$ \scriptsize$(\pm\,1.1)$ & $92.8$ \scriptsize$(\pm\,2.1)$ & $-74.0$ \\
PrAE & $30.0$ \scriptsize$(\pm\,1.1)$ & $63.2$ \scriptsize$(\pm\,2.3)$ & $-33.1$ \\
PredRNet & $31.0$ \scriptsize$(\pm\,1.6)$ & $\underline{95.4}$ \scriptsize$(\pm\,0.4)$ & $-64.4$ \\
RelBase & $36.6$ \scriptsize$(\pm\,0.8)$ & $95.2$ \scriptsize$(\pm\,0.6)$ & $-58.7$ \\
SCL & $\textbf{40.8}$ \scriptsize$(\pm\,3.2)$ & $94.2$ \scriptsize$(\pm\,1.7)$ & $-53.4$ \\
SRAN & $22.7$ \scriptsize$(\pm\,1.1)$ & $84.3$ \scriptsize$(\pm\,0.7)$ & $-61.7$ \\
STSN & $27.3$ \scriptsize$(\pm\,4.6)$ & $84.5$ \scriptsize$(\pm\,12.7)$ & $-57.2$ \\
WReN & $13.5$ \scriptsize$(\pm\,0.1)$ & $43.4$ \scriptsize$(\pm\,1.4)$ & $-29.9$ \\
\bottomrule
\end{tabular}
\caption{\textbf{\texttt{A/ColorSize}.}}
\label{tab:stl-aux-color-size}
\end{table}

\begin{table}[t]
\centering\small
\begin{tabular}{l|ccc}
\toprule
 & Test & Val & Test $-$ Val \\
\midrule
ALANS & $17.7$ \scriptsize$(\pm\,3.2)$ & $20.4$ \scriptsize$(\pm\,6.2)$ & $-\ \ 2.7$ \\
CPCNet & $25.0$ \scriptsize$(\pm\,0.9)$ & $84.4$ \scriptsize$(\pm\,2.7)$ & $-59.4$ \\
CNN-LSTM & $14.7$ \scriptsize$(\pm\,1.7)$ & $53.3$ \scriptsize$(\pm\,5.9)$ & $-38.7$ \\
CoPINet & $17.2$ \scriptsize$(\pm\,0.1)$ & $72.8$ \scriptsize$(\pm\,0.5)$ & $-55.6$ \\
DRNet & $29.5$ \scriptsize$(\pm\,0.5)$ & $\underline{96.0}$ \scriptsize$(\pm\,0.4)$ & $-66.4$ \\
MRNet & $20.0$ \scriptsize$(\pm\,2.6)$ & $91.2$ \scriptsize$(\pm\,4.2)$ & $-71.2$ \\
PrAE & $26.7$ \scriptsize$(\pm\,0.7)$ & $58.8$ \scriptsize$(\pm\,2.6)$ & $-32.1$ \\
PredRNet & $28.0$ \scriptsize$(\pm\,0.7)$ & $91.6$ \scriptsize$(\pm\,0.6)$ & $-63.6$ \\
RelBase & $\underline{29.7}$ \scriptsize$(\pm\,0.6)$ & $93.0$ \scriptsize$(\pm\,4.0)$ & $-63.3$ \\
SCL & $\textbf{32.0}$ \scriptsize$(\pm\,2.3)$ & $\textbf{96.4}$ \scriptsize$(\pm\,0.4)$ & $-64.4$ \\
SRAN & $20.9$ \scriptsize$(\pm\,0.9)$ & $84.4$ \scriptsize$(\pm\,1.9)$ & $-63.5$ \\
STSN & $21.9$ \scriptsize$(\pm\,4.6)$ & $76.2$ \scriptsize$(\pm\,22.2)$ & $-54.3$ \\
WReN & $13.8$ \scriptsize$(\pm\,0.7)$ & $41.3$ \scriptsize$(\pm\,1.0)$ & $-27.5$ \\
\bottomrule
\end{tabular}
\caption{\textbf{\texttt{A/ColorType}.}}
\label{tab:stl-aux-color-type}
\end{table}

\begin{table}[t]
\centering\small
\begin{tabular}{l|ccc}
\toprule
 & Test & Val & Test $-$ Val \\
\midrule
ALANS & $15.7$ \scriptsize$(\pm\,3.2)$ & $21.6$ \scriptsize$(\pm\,11.7)$ & $-\ \ 5.9$ \\
CPCNet & $24.1$ \scriptsize$(\pm\,1.2)$ & $87.0$ \scriptsize$(\pm\,1.8)$ & $-62.9$ \\
CNN-LSTM & $13.0$ \scriptsize$(\pm\,0.1)$ & $53.6$ \scriptsize$(\pm\,0.2)$ & $-40.6$ \\
CoPINet & $19.7$ \scriptsize$(\pm\,0.7)$ & $72.8$ \scriptsize$(\pm\,0.3)$ & $-53.1$ \\
DRNet & $\underline{31.6}$ \scriptsize$(\pm\,1.2)$ & $95.6$ \scriptsize$(\pm\,0.2)$ & $-64.0$ \\
MRNet & $28.2$ \scriptsize$(\pm\,0.9)$ & $\textbf{97.7}$ \scriptsize$(\pm\,0.2)$ & $-69.5$ \\
PrAE & $25.6$ \scriptsize$(\pm\,0.8)$ & $65.0$ \scriptsize$(\pm\,3.7)$ & $-39.4$ \\
PredRNet & $27.9$ \scriptsize$(\pm\,0.5)$ & $94.8$ \scriptsize$(\pm\,0.3)$ & $-66.9$ \\
RelBase & $31.1$ \scriptsize$(\pm\,1.0)$ & $94.6$ \scriptsize$(\pm\,1.4)$ & $-63.5$ \\
SCL & $\textbf{33.5}$ \scriptsize$(\pm\,0.7)$ & $\underline{97.2}$ \scriptsize$(\pm\,0.6)$ & $-63.7$ \\
SRAN & $23.3$ \scriptsize$(\pm\,0.3)$ & $88.4$ \scriptsize$(\pm\,0.3)$ & $-65.1$ \\
STSN & $12.3$ \scriptsize$(\pm\,0.1)$ & $12.5$ \scriptsize$(\pm\,0.5)$ & $-\ \ 0.2$ \\
WReN & $14.1$ \scriptsize$(\pm\,0.2)$ & $50.9$ \scriptsize$(\pm\,0.7)$ & $-36.8$ \\
\bottomrule
\end{tabular}
\caption{\textbf{\texttt{A/SizeType}.}}
\label{tab:stl-aux-size-type}
\end{table}

\begin{table}[t]
\centering\small
\begin{tabular}{l|ccc}
\toprule
 & Test & Val & Test $-$ Val \\
\midrule
ALANS & $24.8$ \scriptsize$(\pm\,18.8)$ & $26.0$ \scriptsize$(\pm\,20.2)$ & $-\ \ 1.2$ \\
CPCNet & $50.5$ \scriptsize$(\pm\,0.6)$ & $75.5$ \scriptsize$(\pm\,0.7)$ & $-25.0$ \\
CNN-LSTM & $17.2$ \scriptsize$(\pm\,1.5)$ & $29.6$ \scriptsize$(\pm\,2.7)$ & $-12.4$ \\
CoPINet & $35.8$ \scriptsize$(\pm\,0.6)$ & $48.9$ \scriptsize$(\pm\,0.5)$ & $-13.1$ \\
DRNet & $72.8$ \scriptsize$(\pm\,1.3)$ & $\textbf{96.1}$ \scriptsize$(\pm\,0.5)$ & $-23.3$ \\
MRNet & $34.4$ \scriptsize$(\pm\,3.4)$ & $93.0$ \scriptsize$(\pm\,2.7)$ & $-58.6$ \\
PrAE & $62.3$ \scriptsize$(\pm\,0.9)$ & $65.1$ \scriptsize$(\pm\,3.7)$ & $-\ \ 2.8$ \\
PredRNet & $62.3$ \scriptsize$(\pm\,2.2)$ & $91.0$ \scriptsize$(\pm\,2.5)$ & $-28.7$ \\
RelBase & $\underline{73.0}$ \scriptsize$(\pm\,1.8)$ & $\underline{95.0}$ \scriptsize$(\pm\,0.8)$ & $-22.0$ \\
SCL & $\textbf{75.6}$ \scriptsize$(\pm\,10.1)$ & $88.8$ \scriptsize$(\pm\,7.1)$ & $-13.2$ \\
SRAN & $42.1$ \scriptsize$(\pm\,2.3)$ & $64.4$ \scriptsize$(\pm\,1.8)$ & $-22.4$ \\
STSN & $39.9$ \scriptsize$(\pm\,14.7)$ & $70.7$ \scriptsize$(\pm\,28.9)$ & $-30.8$ \\
WReN & $18.0$ \scriptsize$(\pm\,0.4)$ & $18.8$ \scriptsize$(\pm\,0.4)$ & $-\ \ 0.7$ \\
\bottomrule
\end{tabular}
\caption{\textbf{\texttt{A/Color-Progression}.}}
\label{tab:stl-aux-color-progression}
\end{table}

\begin{table}[t]
\centering\small
\begin{tabular}{l|ccc}
\toprule
 & Test & Val & Test $-$ Val \\
\midrule
ALANS & $18.3$ \scriptsize$(\pm\,6.6)$ & $19.2$ \scriptsize$(\pm\,6.5)$ & $-\ \ 0.9$ \\
CPCNet & $45.9$ \scriptsize$(\pm\,2.7)$ & $74.3$ \scriptsize$(\pm\,1.8)$ & $-28.4$ \\
CNN-LSTM & $17.1$ \scriptsize$(\pm\,3.7)$ & $26.0$ \scriptsize$(\pm\,4.5)$ & $-\ \ 8.9$ \\
CoPINet & $35.2$ \scriptsize$(\pm\,0.5)$ & $45.5$ \scriptsize$(\pm\,0.8)$ & $-10.4$ \\
DRNet & $\textbf{66.7}$ \scriptsize$(\pm\,1.2)$ & $\textbf{93.5}$ \scriptsize$(\pm\,0.5)$ & $-26.8$ \\
MRNet & $35.7$ \scriptsize$(\pm\,5.9)$ & $88.8$ \scriptsize$(\pm\,5.6)$ & $-53.1$ \\
PrAE & $43.0$ \scriptsize$(\pm\,26.5)$ & $43.4$ \scriptsize$(\pm\,26.7)$ & $-\ \ 0.4$ \\
PredRNet & $56.9$ \scriptsize$(\pm\,1.4)$ & $89.5$ \scriptsize$(\pm\,0.4)$ & $-32.6$ \\
RelBase & $\underline{66.2}$ \scriptsize$(\pm\,1.0)$ & $93.5$ \scriptsize$(\pm\,0.2)$ & $-27.3$ \\
SCL & $60.0$ \scriptsize$(\pm\,4.1)$ & $85.0$ \scriptsize$(\pm\,1.5)$ & $-25.1$ \\
SRAN & $39.9$ \scriptsize$(\pm\,2.7)$ & $61.7$ \scriptsize$(\pm\,3.2)$ & $-21.8$ \\
STSN & $25.7$ \scriptsize$(\pm\,10.6)$ & $39.5$ \scriptsize$(\pm\,25.8)$ & $-13.7$ \\
WReN & $17.1$ \scriptsize$(\pm\,0.2)$ & $18.2$ \scriptsize$(\pm\,0.3)$ & $-\ \ 1.1$ \\
\bottomrule
\end{tabular}
\caption{\textbf{\texttt{A/Color-Arithmetic}.}}
\label{tab:stl-aux-color-arithmetic}
\end{table}

\begin{table}[t]
\centering\small
\begin{tabular}{l|ccc}
\toprule
 & Test & Val & Test $-$ Val \\
\midrule
ALANS & $22.4$ \scriptsize$(\pm\,7.7)$ & $21.5$ \scriptsize$(\pm\,7.0)$ & $+\ \ 0.9$ \\
CPCNet & $37.8$ \scriptsize$(\pm\,0.9)$ & $74.5$ \scriptsize$(\pm\,0.5)$ & $-36.7$ \\
CNN-LSTM & $20.6$ \scriptsize$(\pm\,6.7)$ & $24.9$ \scriptsize$(\pm\,0.5)$ & $-\ \ 4.3$ \\
CoPINet & $26.9$ \scriptsize$(\pm\,0.5)$ & $48.7$ \scriptsize$(\pm\,0.5)$ & $-21.8$ \\
DRNet & $63.2$ \scriptsize$(\pm\,0.3)$ & $\textbf{95.1}$ \scriptsize$(\pm\,0.2)$ & $-31.9$ \\
MRNet & $18.6$ \scriptsize$(\pm\,0.1)$ & $92.6$ \scriptsize$(\pm\,0.4)$ & $-74.0$ \\
PrAE & $55.1$ \scriptsize$(\pm\,0.8)$ & $63.4$ \scriptsize$(\pm\,3.7)$ & $-\ \ 8.3$ \\
PredRNet & $48.5$ \scriptsize$(\pm\,0.9)$ & $89.7$ \scriptsize$(\pm\,0.9)$ & $-41.2$ \\
RelBase & $\textbf{65.7}$ \scriptsize$(\pm\,4.6)$ & $\underline{93.7}$ \scriptsize$(\pm\,1.6)$ & $-28.0$ \\
SCL & $\underline{63.9}$ \scriptsize$(\pm\,4.3)$ & $83.4$ \scriptsize$(\pm\,1.4)$ & $-19.5$ \\
SRAN & $34.6$ \scriptsize$(\pm\,3.6)$ & $64.7$ \scriptsize$(\pm\,2.3)$ & $-30.1$ \\
STSN & $20.7$ \scriptsize$(\pm\,7.7)$ & $53.5$ \scriptsize$(\pm\,23.2)$ & $-32.8$ \\
WReN & $17.7$ \scriptsize$(\pm\,0.6)$ & $18.3$ \scriptsize$(\pm\,0.6)$ & $-\ \ 0.6$ \\
\bottomrule
\end{tabular}
\caption{\textbf{\texttt{A/Color-DistributeThree}.}}
\label{tab:stl-aux-color-distributethree}
\end{table}

\FloatBarrier


\addtolength{\tabcolsep}{-2pt}

\begin{table*}[t]
\centering\small
\begin{tabular}{l|cccccccc}
\toprule
 & Mean & \texttt{Center} & \texttt{2x2Grid} & \texttt{3x3Grid} & \texttt{L-R} & \texttt{U-D} & \texttt{O-IC} & \texttt{O-IG} \\
\midrule
ALANS & $27.0$ \scriptsize$(\pm\,8.4)$ & $28.8$ \scriptsize$(\pm\,9.2)$ & $25.6$ \scriptsize$(\pm\,5.2)$ & $26.7$ \scriptsize$(\pm\,7.7)$ & $32.1$ \scriptsize$(\pm\,13.2)$ & $31.9$ \scriptsize$(\pm\,12.0)$ & $24.3$ \scriptsize$(\pm\,8.5)$ & $19.8$ \scriptsize$(\pm\,3.8)$ \\
CPCNet & $70.4$ \scriptsize$(\pm\,6.4)$ & $85.0$ \scriptsize$(\pm\,10.9)$ & $53.1$ \scriptsize$(\pm\,10.9)$ & $45.2$ \scriptsize$(\pm\,7.0)$ & $89.3$ \scriptsize$(\pm\,4.4)$ & $89.9$ \scriptsize$(\pm\,5.1)$ & $84.3$ \scriptsize$(\pm\,1.3)$ & $46.0$ \scriptsize$(\pm\,5.7)$ \\
CNN-LSTM & $27.5$ \scriptsize$(\pm\,1.5)$ & $41.2$ \scriptsize$(\pm\,4.3)$ & $27.1$ \scriptsize$(\pm\,3.1)$ & $24.8$ \scriptsize$(\pm\,3.3)$ & $23.4$ \scriptsize$(\pm\,1.9)$ & $22.5$ \scriptsize$(\pm\,0.2)$ & $28.3$ \scriptsize$(\pm\,2.0)$ & $25.6$ \scriptsize$(\pm\,0.7)$ \\
CoPINet & $43.2$ \scriptsize$(\pm\,0.1)$ & $51.8$ \scriptsize$(\pm\,0.7)$ & $34.0$ \scriptsize$(\pm\,0.6)$ & $29.9$ \scriptsize$(\pm\,0.6)$ & $47.2$ \scriptsize$(\pm\,0.6)$ & $49.9$ \scriptsize$(\pm\,0.6)$ & $49.2$ \scriptsize$(\pm\,0.9)$ & $40.3$ \scriptsize$(\pm\,1.2)$ \\
DRNet & $\textbf{90.9}$ \scriptsize$(\pm\,1.1)$ & $\underline{99.3}$ \scriptsize$(\pm\,0.2)$ & $\underline{91.2}$ \scriptsize$(\pm\,1.5)$ & $\textbf{83.6}$ \scriptsize$(\pm\,1.1)$ & $95.8$ \scriptsize$(\pm\,0.8)$ & $\textbf{98.2}$ \scriptsize$(\pm\,0.4)$ & $\textbf{98.1}$ \scriptsize$(\pm\,0.7)$ & $\textbf{70.3}$ \scriptsize$(\pm\,3.8)$ \\
MRNet & $86.7$ \scriptsize$(\pm\,2.3)$ & $\textbf{99.8}$ \scriptsize$(\pm\,0.1)$ & $82.2$ \scriptsize$(\pm\,10.0)$ & $72.3$ \scriptsize$(\pm\,11.1)$ & $\underline{97.8}$ \scriptsize$(\pm\,2.0)$ & $\underline{97.9}$ \scriptsize$(\pm\,1.3)$ & $94.4$ \scriptsize$(\pm\,3.0)$ & $62.9$ \scriptsize$(\pm\,0.8)$ \\
\bottomrule
\end{tabular}
\caption{\textbf{I-RAVEN.} Results from Table~\ref{tab:attributeless} extended to each matrix configuration.}
\label{tab:stl-configurations-i-raven}
\end{table*}

\begin{table*}[t]
\centering\small
\begin{tabular}{l|cccccccc}
\toprule
 & Mean & \texttt{Center} & \texttt{2x2Grid} & \texttt{3x3Grid} & \texttt{L-R} & \texttt{U-D} & \texttt{O-IC} & \texttt{O-IG} \\
\midrule
ALANS & $15.9$ \scriptsize$(\pm\,2.6)$ & $16.3$ \scriptsize$(\pm\,2.8)$ & $15.8$ \scriptsize$(\pm\,2.9)$ & $16.1$ \scriptsize$(\pm\,2.7)$ & $16.0$ \scriptsize$(\pm\,3.5)$ & $17.1$ \scriptsize$(\pm\,3.6)$ & $15.6$ \scriptsize$(\pm\,1.9)$ & $14.2$ \scriptsize$(\pm\,0.6)$ \\
CPCNet & $66.6$ \scriptsize$(\pm\,5.1)$ & $73.2$ \scriptsize$(\pm\,11.6)$ & $53.4$ \scriptsize$(\pm\,5.0)$ & $50.1$ \scriptsize$(\pm\,2.9)$ & $78.4$ \scriptsize$(\pm\,3.9)$ & $76.9$ \scriptsize$(\pm\,4.4)$ & $74.9$ \scriptsize$(\pm\,3.2)$ & $58.9$ \scriptsize$(\pm\,5.4)$ \\
CNN-LSTM & $28.9$ \scriptsize$(\pm\,0.4)$ & $30.5$ \scriptsize$(\pm\,0.9)$ & $27.4$ \scriptsize$(\pm\,0.7)$ & $28.4$ \scriptsize$(\pm\,0.5)$ & $28.9$ \scriptsize$(\pm\,0.5)$ & $29.0$ \scriptsize$(\pm\,0.8)$ & $28.8$ \scriptsize$(\pm\,0.3)$ & $29.1$ \scriptsize$(\pm\,1.5)$ \\
CoPINet & $41.1$ \scriptsize$(\pm\,0.3)$ & $41.5$ \scriptsize$(\pm\,0.3)$ & $38.2$ \scriptsize$(\pm\,0.3)$ & $34.6$ \scriptsize$(\pm\,0.2)$ & $42.2$ \scriptsize$(\pm\,1.7)$ & $42.4$ \scriptsize$(\pm\,0.2)$ & $46.0$ \scriptsize$(\pm\,0.3)$ & $42.9$ \scriptsize$(\pm\,0.5)$ \\
DRNet & $\underline{83.9}$ \scriptsize$(\pm\,2.7)$ & $\textbf{93.7}$ \scriptsize$(\pm\,1.0)$ & $77.5$ \scriptsize$(\pm\,6.6)$ & $71.4$ \scriptsize$(\pm\,4.2)$ & $88.2$ \scriptsize$(\pm\,0.8)$ & $\underline{90.9}$ \scriptsize$(\pm\,1.3)$ & $\underline{89.6}$ \scriptsize$(\pm\,0.9)$ & $\textbf{76.0}$ \scriptsize$(\pm\,5.3)$ \\
MRNet & $79.5$ \scriptsize$(\pm\,2.0)$ & $\underline{93.2}$ \scriptsize$(\pm\,4.9)$ & $65.4$ \scriptsize$(\pm\,6.2)$ & $59.4$ \scriptsize$(\pm\,5.1)$ & $\textbf{90.1}$ \scriptsize$(\pm\,4.2)$ & $\textbf{91.2}$ \scriptsize$(\pm\,3.8)$ & $89.5$ \scriptsize$(\pm\,1.6)$ & $67.6$ \scriptsize$(\pm\,1.7)$ \\
\bottomrule
\end{tabular}
\caption{\textbf{I-RAVEN-Mesh.} Results from Table~\ref{tab:attributeless} extended to each matrix configuration.}
\label{tab:stl-configurations-mesh}
\end{table*}

\begin{table*}[t]
\centering\small
\begin{tabular}{l|cccccccc}
\toprule
 & Mean & \texttt{Center} & \texttt{2x2Grid} & \texttt{3x3Grid} & \texttt{L-R} & \texttt{U-D} & \texttt{O-IC} & \texttt{O-IG} \\
\midrule
ALANS & $15.2$ \scriptsize$(\pm\,1.4)$ & $14.7$ \scriptsize$(\pm\,2.3)$ & $16.7$ \scriptsize$(\pm\,2.0)$ & $15.1$ \scriptsize$(\pm\,1.5)$ & $15.6$ \scriptsize$(\pm\,2.8)$ & $15.7$ \scriptsize$(\pm\,2.0)$ & $14.6$ \scriptsize$(\pm\,1.8)$ & $14.0$ \scriptsize$(\pm\,1.5)$ \\
CPCNet & $51.2$ \scriptsize$(\pm\,3.8)$ & $48.3$ \scriptsize$(\pm\,6.8)$ & $41.1$ \scriptsize$(\pm\,6.7)$ & $38.9$ \scriptsize$(\pm\,2.6)$ & $57.4$ \scriptsize$(\pm\,2.7)$ & $57.5$ \scriptsize$(\pm\,2.0)$ & $67.1$ \scriptsize$(\pm\,2.4)$ & $48.0$ \scriptsize$(\pm\,5.7)$ \\
CNN-LSTM & $17.0$ \scriptsize$(\pm\,3.1)$ & $17.9$ \scriptsize$(\pm\,4.6)$ & $17.3$ \scriptsize$(\pm\,2.7)$ & $16.3$ \scriptsize$(\pm\,2.3)$ & $15.8$ \scriptsize$(\pm\,1.8)$ & $15.9$ \scriptsize$(\pm\,2.6)$ & $18.3$ \scriptsize$(\pm\,4.6)$ & $17.9$ \scriptsize$(\pm\,3.6)$ \\
CoPINet & $32.5$ \scriptsize$(\pm\,0.2)$ & $33.0$ \scriptsize$(\pm\,0.8)$ & $28.9$ \scriptsize$(\pm\,0.7)$ & $27.0$ \scriptsize$(\pm\,0.2)$ & $30.0$ \scriptsize$(\pm\,0.4)$ & $30.7$ \scriptsize$(\pm\,0.7)$ & $39.2$ \scriptsize$(\pm\,0.6)$ & $38.7$ \scriptsize$(\pm\,1.7)$ \\
DRNet & $\textbf{70.0}$ \scriptsize$(\pm\,1.6)$ & $\underline{66.9}$ \scriptsize$(\pm\,4.0)$ & $65.9$ \scriptsize$(\pm\,1.2)$ & $\textbf{63.9}$ \scriptsize$(\pm\,1.1)$ & $\textbf{68.1}$ \scriptsize$(\pm\,1.7)$ & $\textbf{70.1}$ \scriptsize$(\pm\,3.2)$ & $\textbf{82.7}$ \scriptsize$(\pm\,2.0)$ & $\underline{72.5}$ \scriptsize$(\pm\,1.6)$ \\
MRNet & $33.6$ \scriptsize$(\pm\,8.2)$ & $27.1$ \scriptsize$(\pm\,7.1)$ & $39.6$ \scriptsize$(\pm\,1.4)$ & $39.4$ \scriptsize$(\pm\,3.0)$ & $21.1$ \scriptsize$(\pm\,16.4)$ & $20.3$ \scriptsize$(\pm\,17.3)$ & $39.8$ \scriptsize$(\pm\,16.9)$ & $48.4$ \scriptsize$(\pm\,5.0)$ \\
\bottomrule
\end{tabular}
\caption{\textbf{\texttt{A/Color}.} Results from Table~\ref{tab:attributeless} extended to each matrix configuration.}
\label{tab:stl-configurations-color}
\end{table*}

\begin{table*}[t]
\centering\small
\begin{tabular}{l|cccccccc}
\toprule
 & Mean & \texttt{Center} & \texttt{2x2Grid} & \texttt{3x3Grid} & \texttt{L-R} & \texttt{U-D} & \texttt{O-IC} & \texttt{O-IG} \\
\midrule
ALANS & $16.0$ \scriptsize$(\pm\,1.0)$ & $15.0$ \scriptsize$(\pm\,1.9)$ & $18.1$ \scriptsize$(\pm\,0.8)$ & $16.8$ \scriptsize$(\pm\,1.3)$ & $16.7$ \scriptsize$(\pm\,1.6)$ & $16.5$ \scriptsize$(\pm\,1.0)$ & $15.2$ \scriptsize$(\pm\,0.5)$ & $14.0$ \scriptsize$(\pm\,2.6)$ \\
CPCNet & $68.3$ \scriptsize$(\pm\,4.0)$ & $89.4$ \scriptsize$(\pm\,8.8)$ & $29.1$ \scriptsize$(\pm\,2.3)$ & $30.1$ \scriptsize$(\pm\,3.6)$ & $96.1$ \scriptsize$(\pm\,3.1)$ & $96.5$ \scriptsize$(\pm\,2.8)$ & $94.3$ \scriptsize$(\pm\,2.6)$ & $42.6$ \scriptsize$(\pm\,6.2)$ \\
CNN-LSTM & $24.0$ \scriptsize$(\pm\,2.9)$ & $43.8$ \scriptsize$(\pm\,5.6)$ & $13.8$ \scriptsize$(\pm\,1.8)$ & $14.5$ \scriptsize$(\pm\,1.2)$ & $25.4$ \scriptsize$(\pm\,1.9)$ & $26.0$ \scriptsize$(\pm\,2.9)$ & $29.1$ \scriptsize$(\pm\,5.1)$ & $15.1$ \scriptsize$(\pm\,3.2)$ \\
CoPINet & $41.3$ \scriptsize$(\pm\,1.6)$ & $51.2$ \scriptsize$(\pm\,2.6)$ & $22.2$ \scriptsize$(\pm\,0.7)$ & $22.2$ \scriptsize$(\pm\,0.7)$ & $53.1$ \scriptsize$(\pm\,2.4)$ & $52.8$ \scriptsize$(\pm\,2.7)$ & $53.3$ \scriptsize$(\pm\,1.6)$ & $34.5$ \scriptsize$(\pm\,1.6)$ \\
DRNet & $\textbf{77.5}$ \scriptsize$(\pm\,0.9)$ & $\textbf{99.4}$ \scriptsize$(\pm\,0.1)$ & $41.8$ \scriptsize$(\pm\,3.3)$ & $41.3$ \scriptsize$(\pm\,1.6)$ & $97.3$ \scriptsize$(\pm\,0.6)$ & $\underline{98.6}$ \scriptsize$(\pm\,0.3)$ & $\textbf{99.2}$ \scriptsize$(\pm\,0.1)$ & $\textbf{65.4}$ \scriptsize$(\pm\,1.2)$ \\
MRNet & $62.6$ \scriptsize$(\pm\,2.6)$ & $92.0$ \scriptsize$(\pm\,13.7)$ & $19.5$ \scriptsize$(\pm\,3.2)$ & $18.2$ \scriptsize$(\pm\,4.5)$ & $\underline{98.1}$ \scriptsize$(\pm\,2.5)$ & $97.1$ \scriptsize$(\pm\,3.7)$ & $96.4$ \scriptsize$(\pm\,4.1)$ & $16.9$ \scriptsize$(\pm\,0.8)$ \\
\bottomrule
\end{tabular}
\caption{\textbf{\texttt{A/Position}.} Results from Table~\ref{tab:attributeless} extended to each matrix configuration.}
\label{tab:stl-configurations-position}
\end{table*}

\begin{table*}[t]
\centering\small
\begin{tabular}{l|cccccccc}
\toprule
 & Mean & \texttt{Center} & \texttt{2x2Grid} & \texttt{3x3Grid} & \texttt{L-R} & \texttt{U-D} & \texttt{O-IC} & \texttt{O-IG} \\
\midrule
ALANS & $23.3$ \scriptsize$(\pm\,6.5)$ & $23.9$ \scriptsize$(\pm\,7.4)$ & $23.6$ \scriptsize$(\pm\,4.9)$ & $22.1$ \scriptsize$(\pm\,5.8)$ & $26.6$ \scriptsize$(\pm\,8.1)$ & $26.9$ \scriptsize$(\pm\,8.6)$ & $22.0$ \scriptsize$(\pm\,7.1)$ & $18.4$ \scriptsize$(\pm\,3.9)$ \\
CPCNet & $43.5$ \scriptsize$(\pm\,3.5)$ & $48.8$ \scriptsize$(\pm\,6.3)$ & $38.2$ \scriptsize$(\pm\,3.9)$ & $36.3$ \scriptsize$(\pm\,1.8)$ & $51.8$ \scriptsize$(\pm\,4.1)$ & $50.6$ \scriptsize$(\pm\,1.7)$ & $47.8$ \scriptsize$(\pm\,6.2)$ & $31.0$ \scriptsize$(\pm\,2.8)$ \\
CNN-LSTM & $13.6$ \scriptsize$(\pm\,1.4)$ & $15.8$ \scriptsize$(\pm\,2.4)$ & $16.6$ \scriptsize$(\pm\,1.3)$ & $15.0$ \scriptsize$(\pm\,2.3)$ & $12.7$ \scriptsize$(\pm\,1.3)$ & $12.3$ \scriptsize$(\pm\,1.1)$ & $11.2$ \scriptsize$(\pm\,1.9)$ & $11.5$ \scriptsize$(\pm\,1.1)$ \\
CoPINet & $21.8$ \scriptsize$(\pm\,0.2)$ & $22.9$ \scriptsize$(\pm\,1.0)$ & $25.9$ \scriptsize$(\pm\,1.2)$ & $27.0$ \scriptsize$(\pm\,0.3)$ & $20.1$ \scriptsize$(\pm\,0.9)$ & $22.1$ \scriptsize$(\pm\,0.5)$ & $15.0$ \scriptsize$(\pm\,0.5)$ & $19.4$ \scriptsize$(\pm\,0.6)$ \\
DRNet & $\underline{54.3}$ \scriptsize$(\pm\,3.0)$ & $\underline{63.0}$ \scriptsize$(\pm\,4.8)$ & $\underline{56.5}$ \scriptsize$(\pm\,1.3)$ & $52.7$ \scriptsize$(\pm\,1.5)$ & $52.4$ \scriptsize$(\pm\,2.8)$ & $\underline{56.6}$ \scriptsize$(\pm\,3.9)$ & $\underline{48.9}$ \scriptsize$(\pm\,7.5)$ & $\underline{50.2}$ \scriptsize$(\pm\,5.5)$ \\
MRNet & $20.6$ \scriptsize$(\pm\,5.0)$ & $20.8$ \scriptsize$(\pm\,12.1)$ & $37.6$ \scriptsize$(\pm\,6.3)$ & $35.1$ \scriptsize$(\pm\,5.7)$ & $8.1$ \scriptsize$(\pm\,4.1)$ & $8.7$ \scriptsize$(\pm\,2.7)$ & $12.1$ \scriptsize$(\pm\,5.1)$ & $21.7$ \scriptsize$(\pm\,3.1)$ \\
\bottomrule
\end{tabular}
\caption{\textbf{\texttt{A/Size}.} Results from Table~\ref{tab:attributeless} extended to each matrix configuration.}
\label{tab:stl-configurations-size}
\end{table*}

\begin{table*}[t]
\centering\small
\begin{tabular}{l|cccccccc}
\toprule
 & Mean & \texttt{Center} & \texttt{2x2Grid} & \texttt{3x3Grid} & \texttt{L-R} & \texttt{U-D} & \texttt{O-IC} & \texttt{O-IG} \\
\midrule
ALANS & $19.0$ \scriptsize$(\pm\,3.4)$ & $18.5$ \scriptsize$(\pm\,4.9)$ & $19.7$ \scriptsize$(\pm\,1.7)$ & $18.9$ \scriptsize$(\pm\,3.8)$ & $22.1$ \scriptsize$(\pm\,3.5)$ & $20.8$ \scriptsize$(\pm\,4.5)$ & $17.6$ \scriptsize$(\pm\,3.1)$ & $15.2$ \scriptsize$(\pm\,3.1)$ \\
CPCNet & $38.6$ \scriptsize$(\pm\,4.3)$ & $33.3$ \scriptsize$(\pm\,4.5)$ & $45.5$ \scriptsize$(\pm\,5.0)$ & $42.4$ \scriptsize$(\pm\,3.1)$ & $40.8$ \scriptsize$(\pm\,3.4)$ & $40.4$ \scriptsize$(\pm\,4.3)$ & $32.2$ \scriptsize$(\pm\,5.0)$ & $34.7$ \scriptsize$(\pm\,6.6)$ \\
CNN-LSTM & $14.5$ \scriptsize$(\pm\,0.8)$ & $13.1$ \scriptsize$(\pm\,0.7)$ & $17.0$ \scriptsize$(\pm\,0.7)$ & $17.3$ \scriptsize$(\pm\,0.8)$ & $13.8$ \scriptsize$(\pm\,1.9)$ & $13.5$ \scriptsize$(\pm\,0.8)$ & $13.9$ \scriptsize$(\pm\,1.8)$ & $12.9$ \scriptsize$(\pm\,1.1)$ \\
CoPINet & $19.8$ \scriptsize$(\pm\,0.9)$ & $18.4$ \scriptsize$(\pm\,1.8)$ & $25.8$ \scriptsize$(\pm\,0.5)$ & $25.5$ \scriptsize$(\pm\,0.6)$ & $18.7$ \scriptsize$(\pm\,1.7)$ & $16.9$ \scriptsize$(\pm\,1.6)$ & $13.8$ \scriptsize$(\pm\,1.0)$ & $19.5$ \scriptsize$(\pm\,0.8)$ \\
DRNet & $\underline{44.3}$ \scriptsize$(\pm\,0.8)$ & $38.7$ \scriptsize$(\pm\,1.8)$ & $\underline{52.0}$ \scriptsize$(\pm\,1.5)$ & $50.2$ \scriptsize$(\pm\,2.4)$ & $41.6$ \scriptsize$(\pm\,0.4)$ & $41.4$ \scriptsize$(\pm\,0.8)$ & $\underline{38.7}$ \scriptsize$(\pm\,1.7)$ & $\textbf{47.3}$ \scriptsize$(\pm\,0.3)$ \\
MRNet & $19.4$ \scriptsize$(\pm\,0.3)$ & $2.7$ \scriptsize$(\pm\,0.2)$ & $48.0$ \scriptsize$(\pm\,1.8)$ & $44.9$ \scriptsize$(\pm\,0.4)$ & $9.0$ \scriptsize$(\pm\,3.8)$ & $7.0$ \scriptsize$(\pm\,0.9)$ & $6.6$ \scriptsize$(\pm\,1.3)$ & $17.8$ \scriptsize$(\pm\,6.0)$ \\
\bottomrule
\end{tabular}
\caption{\textbf{\texttt{A/Type}.} Results from Table~\ref{tab:attributeless} extended to each matrix configuration.}
\label{tab:stl-configurations-type}
\end{table*}

\begin{table*}[t]
\centering\small
\begin{tabular}{l|cccccccc}
\toprule
 & Mean & \texttt{Center} & \texttt{2x2Grid} & \texttt{3x3Grid} & \texttt{L-R} & \texttt{U-D} & \texttt{O-IC} & \texttt{O-IG} \\
\midrule
ALANS & $15.1$ \scriptsize$(\pm\,3.3)$ & $15.2$ \scriptsize$(\pm\,4.3)$ & $17.2$ \scriptsize$(\pm\,2.7)$ & $17.2$ \scriptsize$(\pm\,3.9)$ & $14.9$ \scriptsize$(\pm\,3.4)$ & $13.8$ \scriptsize$(\pm\,4.5)$ & $13.7$ \scriptsize$(\pm\,1.9)$ & $13.7$ \scriptsize$(\pm\,2.4)$ \\
CPCNet & $33.0$ \scriptsize$(\pm\,5.3)$ & $33.0$ \scriptsize$(\pm\,4.4)$ & $34.6$ \scriptsize$(\pm\,4.7)$ & $32.6$ \scriptsize$(\pm\,2.2)$ & $28.0$ \scriptsize$(\pm\,5.7)$ & $30.8$ \scriptsize$(\pm\,6.5)$ & $35.8$ \scriptsize$(\pm\,7.6)$ & $36.3$ \scriptsize$(\pm\,7.8)$ \\
CNN-LSTM & $13.4$ \scriptsize$(\pm\,0.9)$ & $14.3$ \scriptsize$(\pm\,1.0)$ & $15.6$ \scriptsize$(\pm\,2.4)$ & $14.4$ \scriptsize$(\pm\,1.0)$ & $13.4$ \scriptsize$(\pm\,0.3)$ & $13.8$ \scriptsize$(\pm\,1.1)$ & $11.7$ \scriptsize$(\pm\,1.7)$ & $10.4$ \scriptsize$(\pm\,0.8)$ \\
CoPINet & $18.3$ \scriptsize$(\pm\,0.3)$ & $19.7$ \scriptsize$(\pm\,0.8)$ & $24.8$ \scriptsize$(\pm\,0.3)$ & $23.5$ \scriptsize$(\pm\,2.0)$ & $13.9$ \scriptsize$(\pm\,1.3)$ & $15.2$ \scriptsize$(\pm\,0.3)$ & $12.5$ \scriptsize$(\pm\,0.5)$ & $18.4$ \scriptsize$(\pm\,0.5)$ \\
DRNet & $\underline{38.3}$ \scriptsize$(\pm\,0.5)$ & $\underline{38.3}$ \scriptsize$(\pm\,0.9)$ & $\underline{44.2}$ \scriptsize$(\pm\,2.9)$ & $\underline{39.5}$ \scriptsize$(\pm\,2.2)$ & $\textbf{33.0}$ \scriptsize$(\pm\,1.1)$ & $\underline{35.2}$ \scriptsize$(\pm\,3.3)$ & $\underline{37.2}$ \scriptsize$(\pm\,3.6)$ & $\underline{40.2}$ \scriptsize$(\pm\,3.1)$ \\
MRNet & $18.7$ \scriptsize$(\pm\,1.1)$ & $17.6$ \scriptsize$(\pm\,0.9)$ & $26.5$ \scriptsize$(\pm\,2.6)$ & $26.7$ \scriptsize$(\pm\,1.8)$ & $13.8$ \scriptsize$(\pm\,1.6)$ & $13.9$ \scriptsize$(\pm\,1.1)$ & $14.8$ \scriptsize$(\pm\,1.6)$ & $18.1$ \scriptsize$(\pm\,0.9)$ \\
\bottomrule
\end{tabular}
\caption{\textbf{\texttt{A/ColorSize}.} Results from Table~\ref{tab:attributeless-extended} extended to each matrix configuration.}
\label{tab:stl-configurations-color-size}
\end{table*}

\begin{table*}[t]
\centering\small
\begin{tabular}{l|cccccccc}
\toprule
 & Mean & \texttt{Center} & \texttt{2x2Grid} & \texttt{3x3Grid} & \texttt{L-R} & \texttt{U-D} & \texttt{O-IC} & \texttt{O-IG} \\
\midrule
ALANS & $17.7$ \scriptsize$(\pm\,3.2)$ & $17.4$ \scriptsize$(\pm\,4.4)$ & $18.4$ \scriptsize$(\pm\,1.6)$ & $17.6$ \scriptsize$(\pm\,2.4)$ & $19.7$ \scriptsize$(\pm\,5.9)$ & $19.0$ \scriptsize$(\pm\,5.4)$ & $17.1$ \scriptsize$(\pm\,3.4)$ & $14.9$ \scriptsize$(\pm\,1.3)$ \\
CPCNet & $25.0$ \scriptsize$(\pm\,0.9)$ & $21.0$ \scriptsize$(\pm\,1.2)$ & $33.0$ \scriptsize$(\pm\,4.5)$ & $28.4$ \scriptsize$(\pm\,1.6)$ & $23.3$ \scriptsize$(\pm\,1.1)$ & $20.9$ \scriptsize$(\pm\,1.7)$ & $21.5$ \scriptsize$(\pm\,2.4)$ & $26.5$ \scriptsize$(\pm\,2.0)$ \\
CNN-LSTM & $14.7$ \scriptsize$(\pm\,1.7)$ & $13.9$ \scriptsize$(\pm\,3.2)$ & $18.6$ \scriptsize$(\pm\,1.6)$ & $16.7$ \scriptsize$(\pm\,1.9)$ & $14.7$ \scriptsize$(\pm\,1.3)$ & $12.8$ \scriptsize$(\pm\,0.8)$ & $13.4$ \scriptsize$(\pm\,3.0)$ & $12.4$ \scriptsize$(\pm\,2.8)$ \\
CoPINet & $17.2$ \scriptsize$(\pm\,0.1)$ & $14.4$ \scriptsize$(\pm\,0.5)$ & $23.2$ \scriptsize$(\pm\,0.8)$ & $23.6$ \scriptsize$(\pm\,0.7)$ & $13.6$ \scriptsize$(\pm\,0.7)$ & $13.5$ \scriptsize$(\pm\,0.6)$ & $13.2$ \scriptsize$(\pm\,1.0)$ & $18.6$ \scriptsize$(\pm\,1.0)$ \\
DRNet & $29.5$ \scriptsize$(\pm\,0.5)$ & $22.7$ \scriptsize$(\pm\,2.0)$ & $36.7$ \scriptsize$(\pm\,0.3)$ & $34.7$ \scriptsize$(\pm\,0.5)$ & $24.3$ \scriptsize$(\pm\,1.2)$ & $23.0$ \scriptsize$(\pm\,1.6)$ & $\underline{27.2}$ \scriptsize$(\pm\,0.6)$ & $\textbf{38.0}$ \scriptsize$(\pm\,0.2)$ \\
MRNet & $20.0$ \scriptsize$(\pm\,2.6)$ & $15.5$ \scriptsize$(\pm\,0.6)$ & $31.3$ \scriptsize$(\pm\,2.7)$ & $28.6$ \scriptsize$(\pm\,0.7)$ & $15.0$ \scriptsize$(\pm\,3.9)$ & $14.6$ \scriptsize$(\pm\,4.4)$ & $15.1$ \scriptsize$(\pm\,3.6)$ & $19.7$ \scriptsize$(\pm\,3.7)$ \\
\bottomrule
\end{tabular}
\caption{\textbf{\texttt{A/ColorType}.} Results from Table~\ref{tab:attributeless-extended} extended to each matrix configuration.}
\label{tab:stl-configurations-color-type}
\end{table*}

\begin{table*}[t]
\centering\small
\begin{tabular}{l|cccccccc}
\toprule
 & Mean & \texttt{Center} & \texttt{2x2Grid} & \texttt{3x3Grid} & \texttt{L-R} & \texttt{U-D} & \texttt{O-IC} & \texttt{O-IG} \\
\midrule
ALANS & $15.7$ \scriptsize$(\pm\,3.2)$ & $14.0$ \scriptsize$(\pm\,2.0)$ & $18.6$ \scriptsize$(\pm\,5.4)$ & $18.2$ \scriptsize$(\pm\,5.7)$ & $15.1$ \scriptsize$(\pm\,2.0)$ & $14.8$ \scriptsize$(\pm\,1.8)$ & $14.0$ \scriptsize$(\pm\,1.3)$ & $15.6$ \scriptsize$(\pm\,5.3)$ \\
CPCNet & $24.1$ \scriptsize$(\pm\,1.2)$ & $26.3$ \scriptsize$(\pm\,1.8)$ & $30.1$ \scriptsize$(\pm\,2.0)$ & $29.0$ \scriptsize$(\pm\,0.4)$ & $22.3$ \scriptsize$(\pm\,0.6)$ & $22.8$ \scriptsize$(\pm\,0.5)$ & $17.8$ \scriptsize$(\pm\,1.4)$ & $20.1$ \scriptsize$(\pm\,2.5)$ \\
CNN-LSTM & $13.0$ \scriptsize$(\pm\,0.1)$ & $13.1$ \scriptsize$(\pm\,0.4)$ & $14.2$ \scriptsize$(\pm\,0.9)$ & $14.0$ \scriptsize$(\pm\,0.6)$ & $13.2$ \scriptsize$(\pm\,0.9)$ & $12.1$ \scriptsize$(\pm\,0.0)$ & $12.3$ \scriptsize$(\pm\,0.7)$ & $12.6$ \scriptsize$(\pm\,1.2)$ \\
CoPINet & $19.7$ \scriptsize$(\pm\,0.7)$ & $20.9$ \scriptsize$(\pm\,1.6)$ & $23.5$ \scriptsize$(\pm\,0.8)$ & $22.6$ \scriptsize$(\pm\,0.9)$ & $19.1$ \scriptsize$(\pm\,1.1)$ & $19.0$ \scriptsize$(\pm\,0.7)$ & $17.4$ \scriptsize$(\pm\,1.8)$ & $15.5$ \scriptsize$(\pm\,0.9)$ \\
DRNet & $\underline{31.6}$ \scriptsize$(\pm\,1.2)$ & $\underline{35.6}$ \scriptsize$(\pm\,2.6)$ & $\underline{40.4}$ \scriptsize$(\pm\,0.7)$ & $37.1$ \scriptsize$(\pm\,1.1)$ & $\underline{28.4}$ \scriptsize$(\pm\,3.1)$ & $27.5$ \scriptsize$(\pm\,1.7)$ & $\underline{22.9}$ \scriptsize$(\pm\,0.7)$ & $\textbf{29.4}$ \scriptsize$(\pm\,1.5)$ \\
MRNet & $28.2$ \scriptsize$(\pm\,0.9)$ & $31.7$ \scriptsize$(\pm\,2.0)$ & $34.9$ \scriptsize$(\pm\,2.4)$ & $31.0$ \scriptsize$(\pm\,1.5)$ & $27.2$ \scriptsize$(\pm\,3.5)$ & $27.1$ \scriptsize$(\pm\,3.3)$ & $21.1$ \scriptsize$(\pm\,1.3)$ & $24.4$ \scriptsize$(\pm\,2.1)$ \\
\bottomrule
\end{tabular}
\caption{\textbf{\texttt{A/SizeType}.} Results from Table~\ref{tab:attributeless-extended} extended to each matrix configuration.}
\label{tab:stl-configurations-size-type}
\end{table*}

\begin{table*}[t]
\centering\small
\begin{tabular}{l|cccccccc}
\toprule
 & Mean & \texttt{Center} & \texttt{2x2Grid} & \texttt{3x3Grid} & \texttt{L-R} & \texttt{U-D} & \texttt{O-IC} & \texttt{O-IG} \\
\midrule
ALANS & $24.8$ \scriptsize$(\pm\,18.8)$ & $24.9$ \scriptsize$(\pm\,21.3)$ & $24.5$ \scriptsize$(\pm\,13.8)$ & $24.3$ \scriptsize$(\pm\,16.4)$ & $26.3$ \scriptsize$(\pm\,23.3)$ & $29.3$ \scriptsize$(\pm\,26.3)$ & $24.5$ \scriptsize$(\pm\,20.1)$ & $19.7$ \scriptsize$(\pm\,10.1)$ \\
CPCNet & $50.5$ \scriptsize$(\pm\,0.6)$ & $51.9$ \scriptsize$(\pm\,0.3)$ & $37.6$ \scriptsize$(\pm\,2.6)$ & $33.9$ \scriptsize$(\pm\,1.5)$ & $59.0$ \scriptsize$(\pm\,0.6)$ & $58.9$ \scriptsize$(\pm\,1.5)$ & $67.7$ \scriptsize$(\pm\,2.4)$ & $44.8$ \scriptsize$(\pm\,0.8)$ \\
CNN-LSTM & $17.2$ \scriptsize$(\pm\,1.5)$ & $20.3$ \scriptsize$(\pm\,2.2)$ & $17.7$ \scriptsize$(\pm\,1.4)$ & $16.9$ \scriptsize$(\pm\,2.4)$ & $16.0$ \scriptsize$(\pm\,1.3)$ & $15.6$ \scriptsize$(\pm\,1.4)$ & $18.0$ \scriptsize$(\pm\,1.0)$ & $15.7$ \scriptsize$(\pm\,1.9)$ \\
CoPINet & $35.8$ \scriptsize$(\pm\,0.6)$ & $37.3$ \scriptsize$(\pm\,0.9)$ & $29.8$ \scriptsize$(\pm\,1.8)$ & $28.0$ \scriptsize$(\pm\,0.3)$ & $35.8$ \scriptsize$(\pm\,0.5)$ & $35.8$ \scriptsize$(\pm\,0.9)$ & $44.2$ \scriptsize$(\pm\,0.8)$ & $40.1$ \scriptsize$(\pm\,2.5)$ \\
DRNet & $72.8$ \scriptsize$(\pm\,1.3)$ & $71.3$ \scriptsize$(\pm\,2.5)$ & $71.9$ \scriptsize$(\pm\,1.8)$ & $\underline{65.5}$ \scriptsize$(\pm\,0.7)$ & $66.8$ \scriptsize$(\pm\,1.8)$ & $70.4$ \scriptsize$(\pm\,2.7)$ & $\textbf{85.4}$ \scriptsize$(\pm\,0.7)$ & $\textbf{77.6}$ \scriptsize$(\pm\,1.4)$ \\
MRNet & $34.4$ \scriptsize$(\pm\,3.4)$ & $45.6$ \scriptsize$(\pm\,6.1)$ & $33.1$ \scriptsize$(\pm\,8.9)$ & $30.1$ \scriptsize$(\pm\,9.1)$ & $24.1$ \scriptsize$(\pm\,1.8)$ & $23.7$ \scriptsize$(\pm\,4.5)$ & $49.6$ \scriptsize$(\pm\,1.5)$ & $35.2$ \scriptsize$(\pm\,1.1)$ \\
\bottomrule
\end{tabular}
\caption{\textbf{\texttt{A/Color-Progression}.} Results from Table~\ref{tab:attributeless-extended} extended to each matrix configuration.}
\label{tab:stl-configurations-color-progression}
\end{table*}

\begin{table*}[t]
\centering\small
\begin{tabular}{l|cccccccc}
\toprule
 & Mean & \texttt{Center} & \texttt{2x2Grid} & \texttt{3x3Grid} & \texttt{L-R} & \texttt{U-D} & \texttt{O-IC} & \texttt{O-IG} \\
\midrule
ALANS & $18.3$ \scriptsize$(\pm\,6.6)$ & $18.6$ \scriptsize$(\pm\,6.4)$ & $18.2$ \scriptsize$(\pm\,5.5)$ & $17.0$ \scriptsize$(\pm\,5.0)$ & $20.0$ \scriptsize$(\pm\,8.2)$ & $18.8$ \scriptsize$(\pm\,7.6)$ & $18.1$ \scriptsize$(\pm\,8.0)$ & $17.4$ \scriptsize$(\pm\,5.7)$ \\
CPCNet & $45.9$ \scriptsize$(\pm\,2.7)$ & $41.2$ \scriptsize$(\pm\,3.3)$ & $39.2$ \scriptsize$(\pm\,5.7)$ & $35.6$ \scriptsize$(\pm\,5.2)$ & $50.5$ \scriptsize$(\pm\,1.0)$ & $47.7$ \scriptsize$(\pm\,0.8)$ & $62.2$ \scriptsize$(\pm\,1.7)$ & $45.5$ \scriptsize$(\pm\,2.7)$ \\
CNN-LSTM & $17.1$ \scriptsize$(\pm\,3.7)$ & $18.8$ \scriptsize$(\pm\,5.9)$ & $17.3$ \scriptsize$(\pm\,4.2)$ & $17.3$ \scriptsize$(\pm\,3.4)$ & $15.8$ \scriptsize$(\pm\,3.2)$ & $15.6$ \scriptsize$(\pm\,2.4)$ & $17.2$ \scriptsize$(\pm\,3.9)$ & $17.8$ \scriptsize$(\pm\,3.6)$ \\
CoPINet & $35.2$ \scriptsize$(\pm\,0.5)$ & $35.9$ \scriptsize$(\pm\,0.5)$ & $31.4$ \scriptsize$(\pm\,0.5)$ & $27.7$ \scriptsize$(\pm\,0.7)$ & $34.1$ \scriptsize$(\pm\,1.2)$ & $33.5$ \scriptsize$(\pm\,1.6)$ & $43.4$ \scriptsize$(\pm\,0.7)$ & $40.3$ \scriptsize$(\pm\,0.5)$ \\
DRNet & $\textbf{66.7}$ \scriptsize$(\pm\,1.2)$ & $60.0$ \scriptsize$(\pm\,2.2)$ & $\underline{69.1}$ \scriptsize$(\pm\,1.8)$ & $\textbf{62.5}$ \scriptsize$(\pm\,1.9)$ & $\textbf{63.3}$ \scriptsize$(\pm\,2.5)$ & $\textbf{63.2}$ \scriptsize$(\pm\,4.2)$ & $\textbf{77.5}$ \scriptsize$(\pm\,3.0)$ & $\underline{71.6}$ \scriptsize$(\pm\,2.2)$ \\
MRNet & $35.7$ \scriptsize$(\pm\,5.9)$ & $38.9$ \scriptsize$(\pm\,14.1)$ & $38.8$ \scriptsize$(\pm\,3.1)$ & $32.4$ \scriptsize$(\pm\,3.2)$ & $20.6$ \scriptsize$(\pm\,7.5)$ & $20.4$ \scriptsize$(\pm\,6.9)$ & $46.6$ \scriptsize$(\pm\,7.0)$ & $52.4$ \scriptsize$(\pm\,0.8)$ \\
\bottomrule
\end{tabular}
\caption{\textbf{\texttt{A/Color-Arithmetic}.} Results from Table~\ref{tab:attributeless-extended} extended to each matrix configuration.}
\label{tab:stl-configurations-color-arithmetic}
\end{table*}

\begin{table*}[t]
\centering\small
\begin{tabular}{l|cccccccc}
\toprule
 & Mean & \texttt{Center} & \texttt{2x2Grid} & \texttt{3x3Grid} & \texttt{L-R} & \texttt{U-D} & \texttt{O-IC} & \texttt{O-IG} \\
\midrule
ALANS & $22.4$ \scriptsize$(\pm\,7.7)$ & $20.6$ \scriptsize$(\pm\,7.3)$ & $22.0$ \scriptsize$(\pm\,5.2)$ & $23.2$ \scriptsize$(\pm\,8.1)$ & $23.8$ \scriptsize$(\pm\,9.7)$ & $24.2$ \scriptsize$(\pm\,9.8)$ & $23.9$ \scriptsize$(\pm\,9.8)$ & $19.0$ \scriptsize$(\pm\,4.3)$ \\
CPCNet & $37.8$ \scriptsize$(\pm\,0.9)$ & $31.0$ \scriptsize$(\pm\,1.8)$ & $28.7$ \scriptsize$(\pm\,2.9)$ & $28.6$ \scriptsize$(\pm\,2.4)$ & $41.7$ \scriptsize$(\pm\,0.0)$ & $41.0$ \scriptsize$(\pm\,1.3)$ & $54.0$ \scriptsize$(\pm\,0.8)$ & $39.9$ \scriptsize$(\pm\,0.3)$ \\
CNN-LSTM & $20.6$ \scriptsize$(\pm\,6.7)$ & $24.6$ \scriptsize$(\pm\,10.1)$ & $20.1$ \scriptsize$(\pm\,4.9)$ & $19.2$ \scriptsize$(\pm\,5.2)$ & $18.7$ \scriptsize$(\pm\,5.3)$ & $17.9$ \scriptsize$(\pm\,4.1)$ & $22.5$ \scriptsize$(\pm\,9.2)$ & $21.7$ \scriptsize$(\pm\,7.9)$ \\
CoPINet & $26.9$ \scriptsize$(\pm\,0.5)$ & $25.8$ \scriptsize$(\pm\,1.0)$ & $24.1$ \scriptsize$(\pm\,0.1)$ & $21.8$ \scriptsize$(\pm\,0.7)$ & $22.9$ \scriptsize$(\pm\,1.6)$ & $23.4$ \scriptsize$(\pm\,0.2)$ & $34.4$ \scriptsize$(\pm\,1.2)$ & $35.6$ \scriptsize$(\pm\,1.0)$ \\
DRNet & $63.2$ \scriptsize$(\pm\,0.3)$ & $57.1$ \scriptsize$(\pm\,2.0)$ & $\underline{64.1}$ \scriptsize$(\pm\,1.7)$ & $\textbf{58.2}$ \scriptsize$(\pm\,1.8)$ & $61.1$ \scriptsize$(\pm\,0.9)$ & $61.2$ \scriptsize$(\pm\,2.4)$ & $\underline{72.9}$ \scriptsize$(\pm\,2.0)$ & $\underline{67.9}$ \scriptsize$(\pm\,1.6)$ \\
MRNet & $18.6$ \scriptsize$(\pm\,0.1)$ & $20.9$ \scriptsize$(\pm\,3.4)$ & $27.7$ \scriptsize$(\pm\,3.3)$ & $24.0$ \scriptsize$(\pm\,6.6)$ & $11.9$ \scriptsize$(\pm\,6.1)$ & $13.2$ \scriptsize$(\pm\,2.5)$ & $14.3$ \scriptsize$(\pm\,4.6)$ & $18.3$ \scriptsize$(\pm\,6.6)$ \\
\bottomrule
\end{tabular}
\caption{\textbf{\texttt{A/Color-DistributeThree}.} Results from Table~\ref{tab:attributeless-extended} extended to each matrix configuration.}
\label{tab:stl-configurations-color-distributethree}
\end{table*}

\addtolength{\tabcolsep}{-2pt}

\FloatBarrier

\section{Datasheets for datasets}
\label{sec:datasheets}

In what follows, we provide the description of the introduced datasets following the Datasheets for Datasets template proposed by
\citeauthor{gebru2021datasheets}~\shortcite{gebru2021datasheets}.
\bigskip

\definecolor{darkblue}{RGB}{46,25, 110}

\newcommand{\dssectionheader}[1]{%
   \noindent\framebox[\columnwidth]{%
      {\fontfamily{phv}\selectfont \textbf{\textcolor{darkblue}{#1}}}
   }
}

\newcommand{\dsquestion}[1]{%
    {\noindent \fontfamily{phv}\selectfont\small \textcolor{darkblue}{\textbf{#1}}}
}

\newcommand{\dsquestionex}[2]{%
    {\noindent \fontfamily{phv}\selectfont\small \textcolor{darkblue}{\textbf{#1} #2}}
}

\newcommand{\dsanswer}[1]{%
   {\noindent #1 \medskip}
}

\newcommand\multicollinenumbers{%
 \linenumbers
 \def\makeLineNumber{\docolaction{\makeLineNumberLeft}{}{\makeLineNumberRight}}}



\dssectionheader{Motivation}

\dsquestionex{For what purpose was the dataset created?}{Was there a specific task in mind? Was there a specific gap that needed to be filled? Please provide a description.}

\dsanswer{The datasets were created to study generalization and knowledge transfer abilities of AVR models.}

\dsquestion{Who created this dataset (e.g., which team, research group) and on behalf of which entity (e.g., company, institution, organization)?}

\dsanswer{The datasets were created by Mikołaj Małkiński, Adam Kowalczyk and Jacek Mańdziuk from the Warsaw University of Technology, Warsaw, Poland.}

\dsquestionex{Who funded the creation of the dataset?}{If there is an associated grant, please provide the name of the grantor and the grant name and number.}

\dsanswer{This research was carried out with the support of the Laboratory of Bioinformatics and Computational Genomics and the High Performance Computing Center of the Faculty of Mathematics and Information Science Warsaw University of Technology.
Mikołaj Małkiński was funded by the Warsaw University of Technology within the Excellence Initiative: Research University (IDUB) programme.}

\dsquestion{Any other comments?}

\dsanswer{None.}

\bigskip
\dssectionheader{Composition}

\dsquestionex{What do the instances that comprise the dataset represent (e.g., documents, photos, people, countries)?}{ Are there multiple types of instances (e.g., movies, users, and ratings; people and interactions between them; nodes and edges)? Please provide a description.}

\dsanswer{Each dataset instance represents a single Raven's Progressive Matrix, which is a typical task used in human IQ tests.}

\dsquestion{How many instances are there in total (of each type, if appropriate)?}

\dsanswer{Each regime in A-I-RAVEN as well as the I-RAVEN-Mesh dataset contains $70\,000$ instances. The training, validation, and test splits contain $42\,000$, $14\,000$, and $14\,000$ matrices, resp. All together there are $770\,000$ ($11\times 70\,000$) instances.}

\dsquestionex{Does the dataset contain all possible instances or is it a sample (not necessarily random) of instances from a larger set?}{ If the dataset is a sample, then what is the larger set? Is the sample representative of the larger set (e.g., geographic coverage)? If so, please describe how this representativeness was validated/verified. If it is not representative of the larger set, please describe why not (e.g., to cover a more diverse range of instances, because instances were withheld or unavailable).}

\dsanswer{The datasets contain a fixed number of instances generated with the data generator. Using a fixed seed ensures reproducibility of the generation process. The data generator allows to configure the number of generated samples.}

\dsquestionex{What data does each instance consist of? “Raw” data (e.g., unprocessed text or images) or features?}{In either case, please provide a description.}

\dsanswer{Each RPM instance comprises $16$ images that represent the RPM panels, a corresponding index of the correct answer and a representation of rules that govern the matrix. Section~\ref{sec:method} of the paper provides additional details. Each instance is packaged as a separate file in the \texttt{NPZ} format, which is a widely-used binary format to store compressed NumPy arrays.}

\dsquestionex{Is there a label or target associated with each instance?}{If so, please provide a description.}

\dsanswer{See above.}

\dsquestionex{Is any information missing from individual instances?}{If so, please provide a description, explaining why this information is missing (e.g., because it was unavailable). This does not include intentionally removed information, but might include, e.g., redacted text.}

\dsanswer{There's no missing data.}

\dsquestionex{Are relationships between individual instances made explicit (e.g., users’ movie ratings, social network links)?}{If so, please describe how these relationships are made explicit.}

\dsanswer{There are no relationships between individual instances.}

\dsquestionex{Are there recommended data splits (e.g., training, development/validation, testing)?}{If so, please provide a description of these splits, explaining the rationale behind them.}

\dsanswer{The datasets are split into training, validation and test splits. Each generated instance contains the split name in its filename, e.g., the \texttt{RAVEN\_1\_train.npz} file belongs to the train split.
}

\dsquestionex{Are there any errors, sources of noise, or redundancies in the dataset?}{If so, please provide a description.}

\dsanswer{To the best of our knowledge there are no errors, sources of noise, nor redundancies in the datasets.}

\dsquestionex{Is the dataset self-contained, or does it link to or otherwise rely on external resources (e.g., websites, tweets, other datasets)?}{If it links to or relies on external resources, a) are there guarantees that they will exist, and remain constant, over time; b) are there official archival versions of the complete dataset (i.e., including the external resources as they existed at the time the dataset was created); c) are there any restrictions (e.g., licenses, fees) associated with any of the external resources that might apply to a future user? Please provide descriptions of all external resources and any restrictions associated with them, as well as links or other access points, as appropriate.}

\dsanswer{Both datasets are self-contained.}

\dsquestionex{Does the dataset contain data that might be considered confidential (e.g., data that is protected by legal privilege or by doctor-patient confidentiality, data that includes the content of individuals non-public communications)?}{If so, please provide a description.}

\dsanswer{No.}

\dsquestionex{Does the dataset contain data that, if viewed directly, might be offensive, insulting, threatening, or might otherwise cause anxiety?}{If so, please describe why.}

\dsanswer{No.}

\dsquestionex{Does the dataset relate to people?}{If not, you may skip the remaining questions in this section.}

\dsanswer{No.}

\dsquestionex{Does the dataset identify any subpopulations (e.g., by age, gender)?}{If so, please describe how these subpopulations are identified and provide a description of their respective distributions within the dataset.}

\dsanswer{N/A.}

\dsquestionex{Is it possible to identify individuals (i.e., one or more natural persons), either directly or indirectly (i.e., in combination with other data) from the dataset?}{If so, please describe how.}

\dsanswer{N/A.}

\dsquestionex{Does the dataset contain data that might be considered sensitive in any way (e.g., data that reveals racial or ethnic origins, sexual orientations, religious beliefs, political opinions or union memberships, or locations; financial or health data; biometric or genetic data; forms of government identification, such as social security numbers; criminal history)?}{If so, please provide a description.}

\dsanswer{N/A.}

\dsquestion{Any other comments?}

\dsanswer{None.}

\bigskip
\dssectionheader{Collection Process}

\dsquestionex{How was the data associated with each instance acquired?}{Was the data directly observable (e.g., raw text, movie ratings), reported by subjects (e.g., survey responses), or indirectly inferred/derived from other data (e.g., part-of-speech tags, model-based guesses for age or language)? If data was reported by subjects or indirectly inferred/derived from other data, was the data validated/verified? If so, please describe how.}

\dsanswer{The data was generated with a computer program.}

\dsquestionex{What mechanisms or procedures were used to collect the data (e.g., hardware apparatus or sensor, manual human curation, software program, software API)?}{How were these mechanisms or procedures validated?}

\dsanswer{We extended the data generation code used to create I-RAVEN: \url{https://github.com/husheng12345/SRAN}. A subset of the dataset was reviewed manually to ensure correctness of the generated matrices.}

\dsquestion{If the dataset is a sample from a larger set, what was the sampling strategy (e.g., deterministic, probabilistic with specific sampling probabilities)?}

\dsanswer{The dataset is produced by a generator that creates new RPM instances subject to specified constraints through a pseudo-random process. We use a fixed seed to ensure reproducibility of the generation process.}

\dsquestion{Who was involved in the data collection process (e.g., students, crowdworkers, contractors) and how were they compensated (e.g., how much were crowdworkers paid)?}

\dsanswer{The data generator has been written by the authors of this paper without delegating the work to other individuals.}

\dsquestionex{Over what timeframe was the data collected? Does this timeframe match the creation timeframe of the data associated with the instances (e.g., recent crawl of old news articles)?}{If not, please describe the timeframe in which the data associated with the instances was created.}

\dsanswer{Development of the datasets lasted from January 2022 to September 2024.}

\dsquestionex{Were any ethical review processes conducted (e.g., by an institutional review board)?}{If so, please provide a description of these review processes, including the outcomes, as well as a link or other access point to any supporting documentation.}

\dsanswer{N/A.}

\dsquestionex{Does the dataset relate to people?}{If not, you may skip the remaining questions in this section.}

\dsanswer{No.}

\dsquestion{Did you collect the data from the individuals in question directly, or obtain it via third parties or other sources (e.g., websites)?}

\dsanswer{N/A.}

\dsquestionex{Were the individuals in question notified about the data collection?}{If so, please describe (or show with screenshots or other information) how notice was provided, and provide a link or other access point to, or otherwise reproduce, the exact language of the notification itself.}

\dsanswer{N/A.}

\dsquestionex{Did the individuals in question consent to the collection and use of their data?}{If so, please describe (or show with screenshots or other information) how consent was requested and provided, and provide a link or other access point to, or otherwise reproduce, the exact language to which the individuals consented.}

\dsanswer{N/A.}

\dsquestionex{If consent was obtained, were the consenting individuals provided with a mechanism to revoke their consent in the future or for certain uses?}{If so, please provide a description, as well as a link or other access point to the mechanism (if appropriate).}

\dsanswer{N/A.}

\dsquestionex{Has an analysis of the potential impact of the dataset and its use on data subjects (e.g., a data protection impact analysis) been conducted?}{If so, please provide a description of this analysis, including the outcomes, as well as a link or other access point to any supporting documentation.}

\dsanswer{N/A.}

\dsquestion{Any other comments?}

\dsanswer{We bear all responsibility in case of violation of rights.}

\bigskip
\dssectionheader{Preprocessing/cleaning/labeling}

\dsquestionex{Was any preprocessing/cleaning/labeling of the data done (e.g., discretization or bucketing, tokenization, part-of-speech tagging, SIFT feature extraction, removal of instances, processing of missing values)?}{If so, please provide a description. If not, you may skip the remainder of the questions in this section.}

\dsanswer{The data created by the generator is ready to be used in a model. No preprocessing, cleaning, or labeling is required.}

\dsquestionex{Was the “raw” data saved in addition to the preprocessed/cleaned/labeled data (e.g., to support unanticipated future uses)?}{If so, please provide a link or other access point to the “raw” data.}

\dsanswer{N/A.}

\dsquestionex{Is the software used to preprocess/clean/label the instances available?}{If so, please provide a link or other access point.}

\dsanswer{No specific software is required to preprocess, clean, or label the instances. The released code repository contains the code required to reproduce all experiments from the paper, which can be used as a reference implementation for loading the datasets.}

\dsquestion{Any other comments?}

\dsanswer{None.}

\bigskip
\dssectionheader{Uses}

\dsquestionex{Has the dataset been used for any tasks already?}{If so, please provide a description.}

\dsanswer{The datasets have been used to conduct experiments presented in the paper.}

\dsquestionex{Is there a repository that links to any or all papers or systems that use the dataset?}{If so, please provide a link or other access point.}

\dsanswer{N/A.}

\dsquestion{What (other) tasks could the dataset be used for?}

\dsanswer{The datasets could be used in a multi-task setting to improve abstract reasoning capabilities of computer vision models.}

\dsquestionex{Is there anything about the composition of the dataset or the way it was collected and preprocessed/cleaned/labeled that might impact future uses?}{For example, is there anything that a future user might need to know to avoid uses that could result in unfair treatment of individuals or groups (e.g., stereotyping, quality of service issues) or other undesirable harms (e.g., financial harms, legal risks) If so, please provide a description. Is there anything a future user could do to mitigate these undesirable harms?}

\dsanswer{We do not see any undesirable harms that could apply to future users of the datasets.}

\dsquestionex{Are there tasks for which the dataset should not be used?}{If so, please provide a description.}

\dsanswer{The datasets should not be used in human IQ tests, as they were explicitly designed to assess generalization and knowledge transfer abilities of deep learning models.}

\dsquestion{Any other comments?}

\dsanswer{None.}

\bigskip
\dssectionheader{Distribution}

\dsquestionex{Will the dataset be distributed to third parties outside of the entity (e.g., company, institution, organization) on behalf of which the dataset was created?}{If so, please provide a description.}

\dsanswer{The datasets are publicly available.}

\dsquestionex{How will the dataset will be distributed (e.g., tarball on website, API, GitHub)}{Does the dataset have a digital object identifier (DOI)?}

\dsanswer{The datasets are available at the GitHub repository: \url{https://github.com/mikomel/raven}.
Each dataset has a corresponding DOI:
Attributeless-I-RAVEN: \url{https://doi.org/10.57967/hf/2396},
I-RAVEN-Mesh: \url{https://doi.org/10.57967/hf/2397}.}

\dsquestion{When will the dataset be distributed?}

\dsanswer{The datasets are available since June 2024.}

\dsquestionex{Will the dataset be distributed under a copyright or other intellectual property (IP) license, and/or under applicable terms of use (ToU)?}{If so, please describe this license and/or ToU, and provide a link or other access point to, or otherwise reproduce, any relevant licensing terms or ToU, as well as any fees associated with these restrictions.}

\dsanswer{The code repository is released under the GPL-3.0 license. This follows the license associated with the generators of the base datasets -- RAVEN (\url{https://github.com/WellyZhang/RAVEN}) and I-RAVEN (\url{https://github.com/husheng12345/SRAN}). The datasets introduced in this paper are released under the CC license.}

\dsquestionex{Have any third parties imposed IP-based or other restrictions on the data associated with the instances?}{If so, please describe these restrictions, and provide a link or other access point to, or otherwise reproduce, any relevant licensing terms, as well as any fees associated with these restrictions.}

\dsanswer{No.}

\dsquestionex{Do any export controls or other regulatory restrictions apply to the dataset or to individual instances?}{If so, please describe these restrictions, and provide a link or other access point to, or otherwise reproduce, any supporting documentation.}

\dsanswer{No.}

\dsquestion{Any other comments?}

\dsanswer{None.}

\bigskip
\dssectionheader{Maintenance}

\dsquestion{Who will be supporting/hosting/maintaining the dataset?}

\dsanswer{Mikołaj Małkiński is the maintainer of the code repository.}

\dsquestion{How can the owner/curator/manager of the dataset be contacted (e.g., email address)?}

\dsanswer{Email addresses of the corresponding authors are provided on the tile page. In addition, GitHub Issues can be used to conduct public discussions directly in the code repository.}

\dsquestionex{Is there an erratum?}{If so, please provide a link or other access point.}

\dsanswer{No.}

\dsquestionex{Will the dataset be updated (e.g., to correct labeling errors, add new instances, delete instances)?}{If so, please describe how often, by whom, and how updates will be communicated to users (e.g., mailing list, GitHub)?}

\dsanswer{Future changes will be documented in release notes in the code repository.}

\dsquestionex{If the dataset relates to people, are there applicable limits on the retention of the data associated with the instances (e.g., were individuals in question told that their data would be retained for a fixed period of time and then deleted)?}{If so, please describe these limits and explain how they will be enforced.}

\dsanswer{N/A.}

\dsquestionex{Will older versions of the dataset continue to be supported/hosted/maintained?}{If so, please describe how. If not, please describe how its obsolescence will be communicated to users.}

\dsanswer{Older versions of the dataset will be available in the history of the code repository.}

\dsquestionex{If others want to extend/augment/build on/contribute to the dataset, is there a mechanism for them to do so?}{If so, please provide a description. Will these contributions be validated/verified? If so, please describe how. If not, why not? Is there a process for communicating/distributing these contributions to other users? If so, please provide a description.}

\dsanswer{Contributions are welcome. GitHub Issues of the code repository will be used to communicate between contributors and project maintainers.}

\dsquestion{Any other comments?}

\dsanswer{None.}


\end{document}